\documentclass[pmlr,twocolumn,10pt]{jmlr} 
\usepackage{url} 
\usepackage{booktabs}
\usepackage{siunitx}
\usepackage{textcomp}
\usepackage{float}
\usepackage[switch]{lineno}

\theorembodyfont{\upshape}
\theoremheaderfont{\scshape}
\theorempostheader{:}
\theoremsep{\newline}

 \title[RadPhi-3: Small Language Models for Radiology]{RadPhi-3: Small Language Models for Radiology}

 \author{%
  \Name{Mercy Prasanna Ranjit} \addr Microsoft Research India\Email{meranjit@microsoft.com}\\
  \Name{Shaury Srivastav} \addr Microsoft Research India \Email{t-ssrivastav@microsoft.com}\\
  \Name{Tanuja Ganu} \addr Microsoft Research India \Email{taganu@microsoft.com}\\
 }

\begin{document}

\maketitle

\begin{abstract}
    LLM based copilot assistants are useful in everyday tasks. There is also a proliferation in the exploration of AI assistant use cases to support radiology workflows in a reliable manner. In this work, we present \texttt{RadPhi-3}, a Small Language Model instruction tuned from \texttt{Phi-3-mini-4k-instruct} with 3.8B parameters to assist with various tasks in radiology workflows. While impression summary generation has been the primary task which has been explored in prior works w.r.t radiology reports of Chest X-rays, we also explore other useful tasks like change summary generation comparing the current radiology report and its prior report, section extraction from radiology reports, tagging the reports with various pathologies and tubes, lines or devices present in them etc. In-addition, instruction tuning \texttt{RadPhi-3} involved learning from a credible knowledge source used by radiologists, Radiopaedia.org. \texttt{RadPhi-3} can be used both to give reliable answers for radiology related queries as well as perform useful tasks related to radiology reports. \texttt{RadPhi-3} achieves SOTA results on the RaLEs radiology report generation benchmark. 

\end{abstract}
\BlankLine

\paragraph*{Data and Code Availability}
\label{sec:dataavailability}
We leverage publicly available data for training \texttt{RadPhi-3}. We use a combination of annotations from multiple datasets like Mimic-CXR \citep{Johnson2019}, Medical-Diff VQA \citep{10.1145/3580305.3599819}, ChestImagenome \citep{Wu2021-sc} and CheXpert Plus \citep{chambon2024chexpertplusaugmentinglarge} datasets. The Medical-Diff VQA and ChestImagenome dataset annotations are derived from radiology reports pertaining to Chest X-rays of Mimic-CXR dataset. We are not making the code available at this point. 



\paragraph*{Institutional Review Board (IRB)}
Proposed use of public datasets was reviewed by home institution. Under policy, use of de-identified public datasets is classified as Not Human Subjects Research [per 45§46.102(e)(1)(ii), 45§46.102(e)(5)]. Guidance and data reflection questions are provided to researchers including considerations to support representativeness, transparency and intended use.


\section{Introduction}
\label{sec:intro}

\paragraph{Language Models for Radiology}
AI-enabled copilot assistants are becoming popular in every domain with the increasing capabilities of LLMs to perform different tasks via instruction tuning and RLHF \citep{ouyang2022traininglanguagemodelsfollow}. While general domain LLMs can generalize with reasonable performance to perform tasks specific to radiology reports using in-context learning with few shots \citep{liu-etal-2023-exploring-boundaries}, they still lag behind on nuanced tasks like impression generation, which is the conclusion or summary of the report where the radiologist provides their most likely diagnosis based on the findings from the imaging study and long-tail label prediction \citep{zhang-etal-2023-long} unless sophisticated prompt engineering is involved as noted in earlier works \citep{Ranjit2024-pk}. In addition, the privacy-preserving machine learning requirements of the medical domain necessitate specialized radiology models that do not attribute the model performance to an individual record. Given these requirements, specialized radiology models are still needed, and we can't yet completely rely on general domain LLMs for all radiology-specific tasks.

\paragraph{Advantages of SLMs}
The large size of LLMs are prohibitive for easy training and deployments. The \texttt{Phi} series of SLMs (Small Language Models) \citep{abdin2024phi3technicalreporthighly} is designed to be highly capable and cost-effective, outperforming models of similar or larger sizes across various language, reasoning, coding, and math benchmarks. We use the the \texttt{Phi-3-mini-4k-instruct} model to further instruction tune for radiology question answering and radiology report related tasks. The small size of \texttt{Phi-3-mini-4k-instruct} makes it ideal for easy finetuning for specialised tasks and in-house deployments of models to meet the privacy requirements common in medical settings. The latest addition to the \texttt{Phi-3} series, \texttt{Phi-3.5-mini-4k-instruct} has multi-lingual support, which  encourages direct finetuning for radiology specific tasks  in different languages without translating the radiology reports to English.

\paragraph{Radiology Report Tasks}
Impression summarisation from radiology report findings is a key useful task which has been explored in many prior works. \citep{liu2024radiologygptlargelanguagemodel},\citep{liu2023radiologyllama2bestinclasslargelanguage}. However, there are other useful tasks pertaining to radiology reports like, comparison of radiology reports to its prior report to summarize the changes between them, say a pathology condition has changed or a new device is placed etc. Radiology report segmentation is another useful task that can help extract useful sections from a radiology report 
like extraction of findings, impressions and placement description of tubes and lines. Noise removal from radiology reports is another useful utility task that can aid radiologists in their day-to-day workflows.  We finetune \texttt{Phi-3-mini-4k-instruct} in a single stage setup for these tasks. 


\paragraph{Key Contributions}

\begin{enumerate}
    \item We finetune \texttt{Phi-3-mini-4k-instruct} SLM for radiology question answering using Radiopaedia articles \footnote{\url{https://radiopaedia.org/articles}} across fifteen systems and radiology report tasks pertaining to Chest X-rays. The resulting model \texttt{RadPhi-3} outperforms \texttt{RadPhi-2} \citep{Ranjit2024-pk} in radiology question answering and shows significant improvement in semantic tasks like radiology natural language inference and sentence similarity. 
    
    \item We introduce two new radiology report 
 related tasks in addition to the tasks defined in the \texttt{RadPhi-2} work:  radiology report segmentation to extract different sections of a radiology report and  temporal change summary task to generate change summary of medical observations and support devices given the current report and prior report as inputs.
    
    \item We benchmark \texttt{RadPhi-3} on various tasks and benchmarks: RaLEs  \citep{chaves2023RaLEs}, Mimic-CXR \citep{Johnson2019} and CheXpert Plus \citep{chambon2024chexpertplusaugmentinglarge} for impression prediction, Radgraph2 \citep{Dejl2024-mt} for temporal change summary prediction, Medical diff VQA \citep{10.1145/3580305.3599819} for question answering comprehension, RADNLI \citep{Miura2021-bl} and MEDNLI \citep{Shivade2017-zu} for natural language inference, and CheXpert Plus \citep{chambon2024chexpertplusaugmentinglarge} for the radiology report segmentation task. We also benchmark \texttt{RadPhi-3} on a downstream task of multi-label classification of pathologies using Spanish radiology reports from the Padchest dataset \citep{bustos2019padchestlargechestxray}.

\end{enumerate}

\section{Related Work}
\label{sec:relatedwork}

\paragraph{Instruction tuning for radiology}
The work `Rad-Phi2: Instruction Tuning Phi2 for Radiology' \citep{Ranjit2024-pk} was the first to explore instruction tuning for tasks pertaining to radiology report related workflows. They performed instruction tuning of the \texttt{Phi-2} SLM in two stages, first general domain instruction tuning followed by radiology specific instruction tuning and noted that two-stage instructing tuning is important for good performance on radiology report related tasks quoting that \texttt{Phi-2} was not an instruction-tuned model in the first place. They showed that radiology specific models are important for good performance on nuanced tasks like impression prediction and long tail label prediction from radiology reports. They also had two model versions, \texttt{RadPhi-2-Base} for Radiology Question Answering and \texttt{RadPhi-2} for radiology report related tasks. We extend the work of \texttt{RadPhi-2} with additional tasks and datasets and also update the instruction tuning to be based on \texttt{Phi-3-mini-4k-instruct}. We also do not use the general domain instruction tuning step used in \texttt{RadPhi-2} as \texttt{Phi-3-mini-4k-instruct} is already an instruction tuned model.

\paragraph{Phi-3}
\texttt{Phi-3-mini-4k-instruct} \citep{abdin2024phi3technicalreporthighly} is one of the latest addition to the \texttt{Phi} SLM model series. The main difference from the \texttt{Phi2} model is that instruction tuning with direct preference optimization was added in the \texttt{Phi-3} version. We train using the \texttt{Phi3} version for our radiology specific instruction tuning.

\paragraph{Temporal change summary}
The paper `RadGraph2: Modeling Disease Progression in Radiology Reports via Hierarchical Information Extraction' \citep{pmlr-v219-khanna23a} introduced the Radgraph2 dataset \citep{Dejl2024-mt} that has change tracking annotations w.r.t medical observations and support devices in radiology reports to help develop systems that track disease progression over time. We leverage these annotations to extract sections of radiology reports that capture the changes w.r.t medical observations or support devices and use it to benchmark \texttt{RadPhi-3} on the new task of reporting temporal change summary w.r.t the current radiology report.



  \begin{figure*}[h]
    \centering
  \includegraphics[scale=0.45]{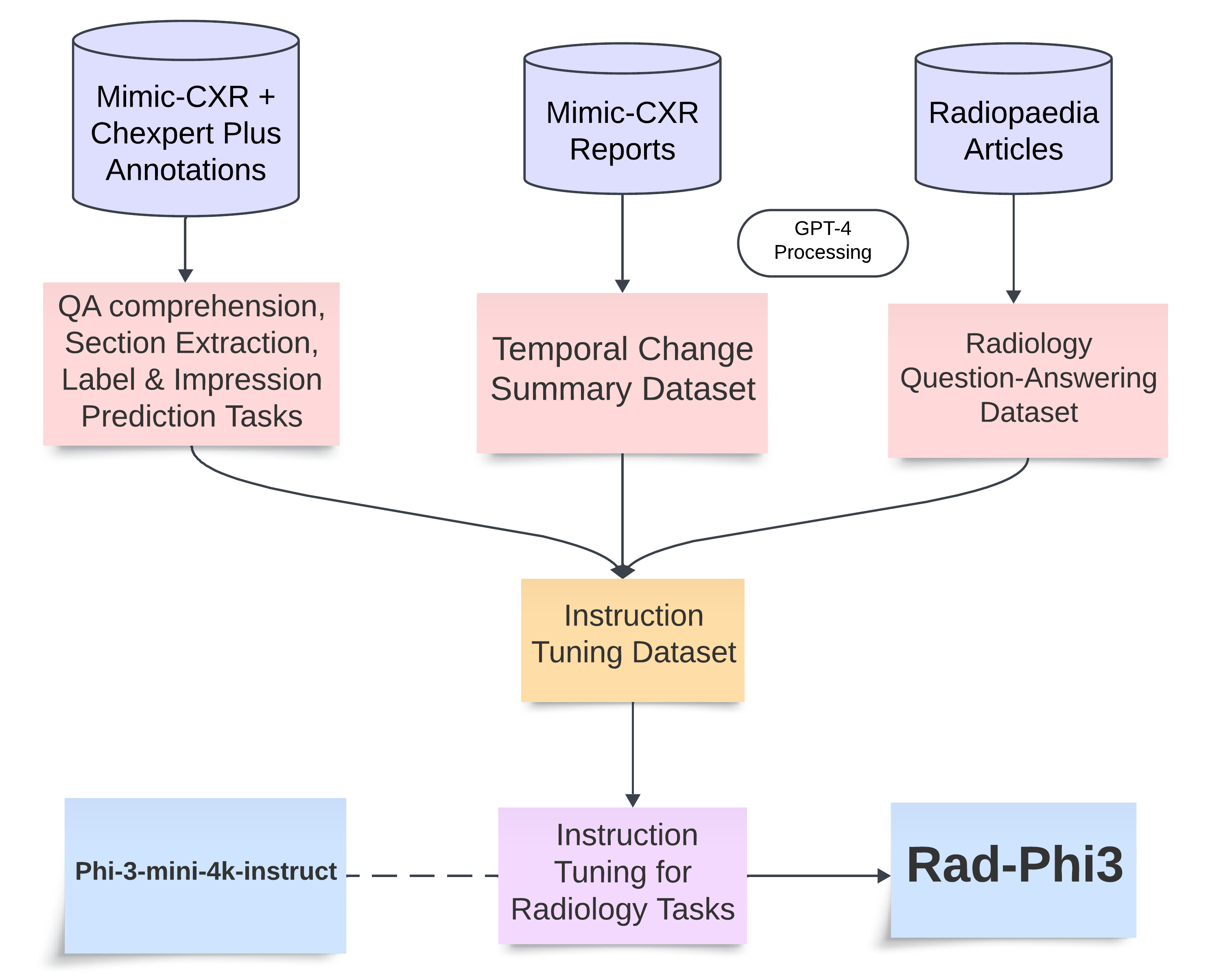}

	\caption{Training Workflow of \texttt{RadPhi-3}.}
	\label{fig:training_workflow}

\end{figure*}

\section{Methodology}
\label{sec:metholdogy}
In this section we explain the training datasets,  finetuning approach, baselines and evaluation metrics we used for instruction tuning \texttt{Phi-3-mini-4k-instruct} for radiology specific tasks.

\subsection{Datasets}
In this section we describe the datasets used for training  \texttt{RadPhi-3}.

\subsubsection{Radiology Question Answering}
\label{sec: radqa}
The Radiology question answering dataset was constructed from Radiopaedia articles pertaining to 15 different systems and contains questions and answers related to radiological appearances of findings, differential diagnosis, assessing prognosis, and suggesting treatments etc. The dataset statistics is available in Table \ref {tab:radiopaedia_stats}. It was constructed with the help of \texttt{GPT-4} by sending the Radiopaedia articles as the context and instructing \texttt{GPT-4} to construct QA pairs using the context. The prompt used for constructing the dataset is available in the appendix section \ref{sec:prompt_radiopaedia}. 

Question-answer pairs generated from the summary articles of Radiopaedia were reserved as the test set. We performed a radiologist evaluation of question-answer pairs generated from 82 randomly sampled articles from the train and test sets, stratified across systems, to check for hallucination of facts, instruction-following hallucinations, and the quality of questions and answers generated. Hallucination of facts refers to errors related to facts in the answers that were not present in the article. Instruction-following hallucinations refer to inadequate adherence to instructions while answering a question. Quality errors refer to question-answer pairs that are not very useful in a clinical setting. The evaluation results are available in Table \ref{tab:radiologist eval_radiopaedia}. There were zero factual hallucinations, five quality-related errors, and one instruction-following hallucination. The examples of these errors are available in Table \ref{tab:example_radiopaedia_qa_error}, \ref{tab:example2_radiopaedia_qa_error}, and \ref{tab:example3_radiopaedia_qa_error}. As seen from these tables, the quality errors were primarily due to inadequate information in the Radiopaedia article.

\subsubsection{Radiology Report Related Tasks}
\label{sec: radiolofytasks}
There are useful tasks pertaining to radiology reports. Examples of such tasks include impression prediction from findings, abnormality and support devices label prediction, QA comprehension, extract findings/impression from a radiology report and cleanup of a noisy radiology report text. We extend the radiology report related instruction tuning dataset in the prior work \texttt{RadPhi-2}  \citep{Ranjit2024-pk} with additional tasks and datasets. This dataset is constructed by using the annotations of the public datasets defined in the data availability section\ref{sec:dataavailability} and translating these annotations to an instruction format using the prompts defined in Table \ref{tab:prompt_design}. We added impressions from the CheXpert Plus dataset for the impression prediction task.

The data split for all the tasks is based on Mimic-CXR's official train, validation and test spilt as all of these datasets are derived from Mimic-CXR annotations. Tasks based on CheXpert Plus uses its official train and validation  splits.  The train, validation and test split for the different tasks is defined in Table \ref{tab:radinstruct_dataset_stats}. 




 We also added the below two new tasks, radiology report generation and temporal change summary to the instruction tuning dataset. 

\subsubsection{Radiology report segmentation} 
\label{sec: reportseg}
CheXpert Plus \citep{chambon2024chexpertplusaugmentinglarge} is a relatively new dataset with 187,711 radiology reports for Chest X-rays, each meticulously divided into subsections defined in Table \ref{tab:chexpert_dataset_stats}. We introduce the radiology report segmentation task leveraging these annotations to segment the radiology reports into different report sections.  We use the official train split for training and validation split for testing. The dataset statistics is available in Table \ref{tab:chexpert_dataset_stats}. The prompt for translating the annotations from CheXpert Plus to the instruction tuning format is available in Table \ref{tab:prompt_design}.

\subsubsection{Temporal change summary} 
\label{sec: tempsumm}
This task is defined as: Given a radiology report and its prior report, extract a change summary of medical observations and support devices. This is extracted under 5 different categories (New, Resolved, Worsened, Stable and Improved) for medical conditions and 4 different categories (New, Removed, Changed and Unchanged) for support devices. 

In addition, we also had a category for extracting the negative findings for medical conditions and recommendations for device adjustments. We use \texttt{GPT-4} to extract the change summary using the prompt defined in the Appendix section \ref{sec:temporal_extraction_summary_prompt}. 
The dataset statistics is available in Table \ref{tab:temporal_dataset_stats}. 

We performed a radiologist evaluation of this dataset by sampling 50 examples from the test set to check for fact hallucinations, hallucinations of categories, and missed findings. Hallucinations of facts refer to errors related to facts in the answers that were not present in the current or prior report. Hallucinations of categories refer to errors where the change summary category (new, improved, worsened, resolved, etc.) was incorrectly assigned. Missed findings errors are related to the missed change mentions. The evaluation results are in Table \ref{tab:radiologist eval_temporal_summary}. There were zero factual hallucinations. There were a few hallucinations with respect to the assignment of change categories, where GPT-4 incorrectly classified the change category. An example of such is available in Table \ref{tab:example_temporal_change_summary_error}.

\subsection{Fine-tuning Details}
As shown in Figure \ref{fig:training_workflow}, we create the instruction tuning dataset from Mimic-CXR reports and related annotations from the datasets described in section \ref{sec: radiolofytasks} for QA comprehension, label prediction, impression prediction, and extraction tasks.

We use \texttt{GPT-4} processing to extract change summaries from Mimic-CXR reports with respect to medical observations and support devices as described in section \ref{sec: tempsumm} and question-answer pairs from radiology articles as described in section \ref{sec: radqa}.

All of these together form the radiology instruction tuning dataset that is used to fine-tune \texttt{RadPhi-3}.

\texttt{RadPhi-3} is fine-tuned from the \texttt{Phi-3-mini-4k-instruct} model checkpoint by continued fine-tuning for 3 epochs. We fine-tune using 4x NVIDIA A100 Tensor Core GPUs with 80GB VRAM using the Huggingface framework with Flash Attention 2, DeepSpeed Stage 3, and the paged\_adamw\_32bit optimizer. A global batch size of 128 was used for 3 epochs, as well as a base learning rate of 5e-5 and a warm-up ratio of 0.1.

\subsection{Baselines}

We use \texttt{RadPhi-2} as our baseline model for both the Radiology QA task and the report related tasks. We also additionally measure the performance of \texttt{Phi-3-mini-4k-instruct} to check the radiology knowledge of a general domain model. We also compare with the existing SOTA approaches depending on the task.

\begin{table*}[!htbp]
\begin{center}
\footnotesize
\caption{Radiology Question Answering Metrics on Radiopaedia Summary Articles.}
\resizebox{\textwidth}{!}{
\begin{tabular}{|lSSSS|}
\hline
\textbf{Models} & \textbf{F1-Score} & \textbf{Precision} & \textbf{Recall} &  \textbf{RougeL}\\
\hline

Phi-3-mini-4k-instruct & 32.91 {\scriptsize \textcolor{gray}{ [31.3, 34.93]}} 
& 21.01 {\scriptsize \textcolor{gray}{[19.33, 22.47]}} & 
\textbf{76.66 {\scriptsize \textcolor{gray}{[73.54, 79.55]}}} &
11.39   {\scriptsize \textcolor{gray}{[9.96, 12.55]}} \\

GPT-4 & 33.98 {\scriptsize \textcolor{gray}{[31.09, 37.02]}} 
& 28.68 {\scriptsize \textcolor{gray}{[26.24, 31.69]}} & 42.07 {\scriptsize \textcolor{gray}{[37.33, 46.96]}} &
23.68 {\scriptsize \textcolor{gray}{[21.87, 26.35]}} \\

RadPhi-2 & 38.98  {\scriptsize \textcolor{gray}{[35.72, 42.83]}} & 40.97 {\scriptsize \textcolor{gray}{[37.35, 46.75]}} & 37.3 {\scriptsize \textcolor{gray}{[32.85, 41.95]}}  & 27.01   {\scriptsize \textcolor{gray}{[23.96, 29.82]}}\\

RadPhi-3 & \textbf{40.33 {\scriptsize \textcolor{gray}{ [36.56, 44.14]}}} & \textbf{42.71 {\scriptsize \textcolor{gray}{[37.71, 46.16]}}} & 38.39 {\scriptsize \textcolor{gray}{[34.92, 43.02]}} & \textbf{28.39  {\scriptsize \textcolor{gray}{ [24.98, 31.54]}}} \\
\hline
\multicolumn{5}{p{430pt}}{\texttt{RadPhi-3} is better than \texttt{RadPhi-2} on all the metrics. \texttt{Phi-3-mini-4k-instruct} has a good recall value indicating the reasonable radiology knowledge of the general domain model.}
\end{tabular}}
\label{tab:radqa_metrics}
\end{center}
\end{table*}

\begin{table*}[!htbp]
\centering
\footnotesize
\caption{Impression Prediction Metrics on Mimic-CXR Test Set}
\resizebox{\textwidth}{!}{
\begin{tabular}{|lcccc|}
\hline
\textbf{Models} & \textbf{F1 Chexbert} & \textbf{F1 RadGraph} & \textbf{RougeL} & \textbf{F1-Score} \\
\hline
Phi-3-mini-4k-instruct & 50.29 {\scriptsize \textcolor{gray}{[50.19, 51.21]}} & 12.80 & 12.18 {\scriptsize \textcolor{gray}{[12.01, 12.48]}} & 24.11  {\scriptsize \textcolor{gray}{[23.56, 24.5]}} \\
GPT-4 (Impression-GPT) & 64.9  &  & 46.0  &  \\
RadPhi-2 & \textbf{67.87 {\scriptsize \textcolor{gray}{[66.95, 68.61]}}} & \textbf{43.06} & \textbf{49.74 {\scriptsize \textcolor{gray}{[49.09, 50.47]}}} & \textbf{59.66  {\scriptsize \textcolor{gray}{[59.15, 60.55]}}} \\
RadPhi-3 & 66.66 {\scriptsize \textcolor{gray}{[65.64, 67.48]}} & 41.74 & 48.33 {\scriptsize \textcolor{gray}{[47.56, 48.91]}} & 58.52 {\scriptsize \textcolor{gray}{[57.66, 59.47]}}\\
\hline
\multicolumn{5}{c}{\texttt{RadPhi-3} performs comparable to \texttt{RadPhi-2} on all the metrics and both are better than Impression GPT.}
\end{tabular}}

\label{tab:impression_prediction_metrics}
\end{table*}

\begin{table*}[!htbp]
\centering
\footnotesize
\caption{Impression Prediction Metrics on ChexPert Plus Validation Set}
\resizebox{\textwidth}{!}{\begin{tabular}{|lcccc|}
\hline
\textbf{Models} & \textbf{F1 Chexbert} & \textbf{F1 RadGraph} & \textbf{RougeL} & \textbf{F1-Score}\\
\hline
Phi-3-mini-4k-instruct & 60.61 {\scriptsize \textcolor{gray}{[56.22, 67.59]}} & 24.92 & 19.87 {\scriptsize \textcolor{gray}{[18.09, 21.74]}} & 33.08 {\scriptsize \textcolor{gray}{[31.32, 34.97]}}\\
RadPhi-2 & 59.4 {\scriptsize \textcolor{gray}{[52.98, 61.59]}} & 33.65 & 40.72 {\scriptsize \textcolor{gray}{[33.95, 44.95]}} & 49.97  {\scriptsize \textcolor{gray}{[44.71, 54.35]}}\\
RadPhi-3 & \textbf{64.02 {\scriptsize \textcolor{gray}{[58.69, 68.92]}}} & \textbf{47.84} & \textbf{47.51 {\scriptsize \textcolor{gray}{[44.81, 49.27]}}} & \textbf{57.89 {\scriptsize \textcolor{gray}{[53.45, 59.35]}}}\\
\bottomrule
\multicolumn{5}{c}{\texttt{RadPhi-3} is better than \texttt{RadPhi-2} in all the metrics.}
\end{tabular}}
\label{tab:impression_prediction_metrics_chexpert}
\end{table*}

\begin{table*}[!htbp]    
\centering
\footnotesize
    \caption{Abnormality and Support Devices Label Metrics - ChestImagenome Labels}
    \resizebox{\textwidth}{!}{  
    \begin{tabular}{|l c c c c|}
        \hline
        \textbf{Task} &\textbf{Metrics} & \textbf{Phi-3-mini-4k-instruct}& \textbf{RadPhi-2}& \textbf{RadPhi-3}\\
        \hline

        Tubes/Lines/Devices Labels&
        \begin{tabular}{l}F1-Score \\ Recall \\ Precision\end{tabular}
        &
        \begin{tabular}{l}
        63.05 {\scriptsize \textcolor{gray}{ [62.72, 63.35]}}\\
        61.43 {\scriptsize \textcolor{gray}{ [61.17, 61.71]}}\\
        71.64 {\scriptsize \textcolor{gray}{ [71.01, 71.95]}}\end{tabular}
        &
        \begin{tabular}{l}
        \textbf{95.54 {\scriptsize \textcolor{gray}{ [95.16, 95.2]}}}\\
        \textbf{95.82 {\scriptsize \textcolor{gray}{ [95.43, 95.51]}}}\\
        \textbf{95.49 {\scriptsize \textcolor{gray}{ [95.12, 95.15]}}}\end{tabular}
        &
        \begin{tabular}{l}
        95.16 {\scriptsize \textcolor{gray}{ [94.54, 95.61]}}\\
        95.49 {\scriptsize \textcolor{gray}{ [94.93, 95.93]}}\\
        95.07 {\scriptsize \textcolor{gray}{ [94.43 95.54]}}\end{tabular}
        \\
        \hline

        Abnormality Labels&
        \begin{tabular}{l}F1-Score \\ Recall \\ Precision\end{tabular}
        &
        \begin{tabular}{l}
        76.72 {\scriptsize \textcolor{gray}{ [76.08, 77.2]}}\\
        76.4 {\scriptsize \textcolor{gray}{ [75.77, 76.82]}}\\
        78.86 {\scriptsize \textcolor{gray}{ [78.18, 79.48]}}\end{tabular}
        &
        \begin{tabular}{l}
        \textbf{94.68 {\scriptsize \textcolor{gray}{ [94.57, 94.87]}}}\\
        \textbf{94.88 {\scriptsize \textcolor{gray}{ [94.82, 95.0]}}}\\
        \textbf{95.54 {\scriptsize \textcolor{gray}{ [95.4, 95.79]}}}\end{tabular}&
        \begin{tabular}{l}
        93.36 {\scriptsize \textcolor{gray}{ [93.04, 93.67]}}\\
        93.89 {\scriptsize \textcolor{gray}{ [93.49, 94.12]}}\\
        93.98 {\scriptsize \textcolor{gray}{[93.59, 94.34]}}\end{tabular}\\
        \hline
\multicolumn{5}{c}{\texttt{RadPhi-2} is slightly better than \texttt{RadPhi-3}. We report the macro F1 score.}
    \end{tabular}}
    \label{tab:label_prediction_metrics}
\end{table*}

\begin{table*}[!htbp]
\centering
\footnotesize
    \caption{QA Comprehension, Findings/Impression Extraction and Clean Radiology Text Task Metrics}
    \resizebox{\textwidth}{!}{\begin{tabular}{|lcccc|}
        \hline
        \textbf{Task} &\textbf{Metrics}&\textbf{Phi-3-mini-4k-instruct} &\textbf{RadPhi2-Instruct}&\textbf{RadPhi-3}\\
        \hline

        QA Comprehension &
        \begin{tabular}{l}F1-Score \\Precision \\Recall \\RougeL \end{tabular}
        &
        \begin{tabular}{l}
        13.07 {\scriptsize \textcolor{gray}{[12.78, 13.24]}}
        \\7.13 {\scriptsize \textcolor{gray}{[7.03, 7.24]}}
        \\78.04 {\scriptsize \textcolor{gray}{[77.45, 78.59]}}
        \\2.96 {\scriptsize \textcolor{gray}{[2.89, 2.98]}}
        \end{tabular}
        &
        \begin{tabular}{l} 
        \textbf{98.1 {\scriptsize \textcolor{gray}{ [97.88, 98.25]}}}\\
        \textbf{97.95 {\scriptsize \textcolor{gray}{ [97.84, 98.15]}}}\\
        \textbf{98.25 {\scriptsize \textcolor{gray}{ [98.13, 98.44]}}}\\
        \textbf{97.61 {\scriptsize \textcolor{gray}{ [97.45, 97.8]}}} 
        \end{tabular}
        &
        \begin{tabular}{l} 
        96.98  {\scriptsize \textcolor{gray}{ [96.88, 97.26]}}\\
        97.05 {\scriptsize \textcolor{gray}{ [96.88, 97.38]}}\\
        96.91 {\scriptsize \textcolor{gray}{ [96.71, 97.21]}}\\
        96.18 {\scriptsize \textcolor{gray}{[95.87, 96.32]}} 
        \end{tabular}
        \\
        \hline
        Cleanup Radiology Text &
        \begin{tabular}{l}
        F1-Score \\
        BLEU-4 \\
        RougeL\\
        AlignScore
        \end{tabular}
        &
        \begin{tabular}{l}
        70.78 {\scriptsize \textcolor{gray}{[70.49, 71.23]}}\\
        17.43 {\scriptsize \textcolor{gray}{[17.08, 17.67]}}\\
        36.32 {\scriptsize \textcolor{gray}{[35.81, 36.61]}}\\
        61.22 {\scriptsize \textcolor{gray}{[60.82, 61.68]}}
        \end{tabular}
        &
        \begin{tabular}{l} 
        92.06 {\scriptsize \textcolor{gray}{[91.97, 92.16]}}\\
        73.23 {\scriptsize \textcolor{gray}{ [72.99, 73.47]}} \\
        86.82 {\scriptsize \textcolor{gray}{ [86.63, 86.95]}} \\
        \textbf{93.97 {\scriptsize \textcolor{gray}{ [93.81, 94.12]}}}
        \end{tabular}
        &
        \begin{tabular}{l} \textbf{92.13 {\scriptsize \textcolor{gray}{ [91.91, 92.38]}}} \\ \textbf{73.29  {\scriptsize \textcolor{gray}{[72.8, 73.74]}}} \\ \textbf{86.95 {\scriptsize \textcolor{gray}{[86.77, 87.14]}}} \\ 93.61 {\scriptsize \textcolor{gray}{[93.5, 93.89]}}\end{tabular}
        \\
        \hline
        Extraction &
         \begin{tabular}{l}F1-Score  \\ BLEU-4 \\RougeL \\ AlignScore \end{tabular}
         &
        \begin{tabular}{l}
        73.93 {\scriptsize \textcolor{gray}{[73.7, 74.12]}}\\
        38.19 {\scriptsize \textcolor{gray}{[37.9, 38.53]}}\\
        62.13 {\scriptsize \textcolor{gray}{[61.85, 62.43]}}\\
        59.83 {\scriptsize \textcolor{gray}{[59.12, 60.44]}}
        \end{tabular}
        &
        \begin{tabular}{l} \textbf{91.85 {\scriptsize \textcolor{gray}{[91.74, 91.99]}}} \\ 72.67 {\scriptsize \textcolor{gray}{[71.99, 73.07}} \\ \textbf{86.55 {\scriptsize \textcolor{gray}{[86.22, 86.83]}}} \\  \textbf{93.75 {\scriptsize \textcolor{gray}{[93.36, 94.02]}}}\end{tabular}
        &
        \begin{tabular}{l} 91.83 {\scriptsize \textcolor{gray}{[91.55, 91.94] }} \\ \textbf{72.72 {\scriptsize \textcolor{gray}{[72.49, 73.21]}}} \\ 86.44 {\scriptsize \textcolor{gray}{[86.3, 86.73]}} \\ 93.35 {\scriptsize \textcolor{gray}{[93.02, 93.57]}}\end{tabular}
        \\
        \hline
        \multicolumn{5}{c}{\texttt{RadPhi-2} performs comparable to \texttt{RadPhi-3} on all the metrics.}
    \end{tabular}} 
    \label{tab:radinstruct_metrics}
\end{table*}

\begin{table*}[!ht]
\centering
\footnotesize
    \caption{NLI Metrics (RADLI + MEDNLI)}
    \begin{tabular}{|lccc|}
    \hline
        \textbf{Model} & \textbf{F1-Score} & \textbf{Precision} & \textbf{Recall} \\
        \hline
        Phi-3-mini-3k-instruct & 22.29 {\scriptsize \textcolor{gray}{[21.59, 30.88]}} & 42.36 {\scriptsize \textcolor{gray}{[40.45, 58.95]}} & 30.04 {\scriptsize \textcolor{gray}{[29.61, 40.93]}}\\
        RadPhi-2 & 0.17 {\scriptsize \textcolor{gray}{[0.02, 0.31]}}& 50.0 {\scriptsize \textcolor{gray}{[5.62, 50.0]}}& 0.09 {\scriptsize \textcolor{gray}{[0.01, 0.16]}}\\
        RadPhi-3 & \textbf{83.89 {\scriptsize \textcolor{gray}{[82.3, 84.56]}}} & \textbf{83.66 {\scriptsize \textcolor{gray}{[82.01, 84.33]}}} & \textbf{84.55 {\scriptsize \textcolor{gray}{[83.05, 85.27]}}}\\
        \hline
        
\multicolumn{4}{c}{\texttt{RadPhi-3} performs significantly better than \texttt{RadPhi-2} on NLI task.}
    \end{tabular}   
    \label{tab:radnli_metrics}
\end{table*}

\subsection{Evaluation Metrics}
\subsubsection {Lexical metrics} We report the ROUGE-L \citep{lin-2004-rouge},  BLEU-4 \citep{papineni-etal-2002-bleu} and F1-score implementation of LlavaMed \citep{li2023llavamedtraininglargelanguageandvision} for the lexical metrics. For label prediction tasks, as its a multi-label classification, we use the f1-score implementation of scikit-learn. We report the macro average. \footnote{\url{https://scikit-learn.org/stable/modules/generated/sklearn.metrics.f1_score.html}}

\subsubsection{Hallucination Metrics} We compute the AlignScore metric \citep{zha2023alignscoreevaluatingfactualconsistency} for all the extraction tasks and the radiology question-answering related task to measure the factual consistency of the outputs w.r.t the inputs. This measures if the model generated text is factually consistent with the input context. 

\subsubsection{Clinical Metrics} We report the F1-CheXbert\citep{smit-etal-2020-combining}, F1-RadGraph \citep{jain2021radgraphextractingclinicalentities} for the clinical metrics.   

F1-RadGraph is based on the RadGraph model \citep{jain2021radgraphextractingclinicalentities} that parses the radiology report into graph of clinical entities (like anatomy and observations and relations) and the metric measures the overlap in clinical entities between the ground truth and candidate report. We use the Radgraph python package(v0.1.2) \footnote{\url{https://pypi.org/project/radgraph/}} which is an implementation of the Radgraph F1 metric based on rewards \citep{jain2021radgraphextractingclinicalentities}. We use the `partial' reward level for computation of this metric. 

F1-CheXbert\citep{smit-etal-2020-combining} measures the micro-averaged F1 score of 14 disease mentions extracted from a generated summary in comparison to the ground truth. We also use the Chexbert python package(v0.0.2) for calculating the F1-Chexbert metric. \footnote{\url{https://pypi.org/project/f1chexbert/}}.

We derive bootstrap confidence intervals for our evaluations by creating 10 resamples with replacement from the test dataset as our dataset is huge, where each resample is the size of the original test set. The temporal summary extraction and radiology report segmentation tasks were an exception to bootstrapping due to the low support for certain categories and we report stratified metrics for these two tasks.

\section{Results}
In this section we discuss the evaluation  results of \texttt{RadPhi-3} on various datasets and benchmarks. 

\subsection{Radiology Question Answering}

The results of the radiology question answering task evaluation on the test set of Radiopaedia summary articles is available in Table \ref{tab:radqa_metrics} and the stratified metrics by system in Table \ref{tab:radiopaedia_metrics} . 

The stratified metrics by system shows that \texttt{RadPhi3-Instruct} performs better than \texttt{GPT-4} on all the systems and by a large margin on systems like Hepatobiliary, Urogenital, Breast, Musculoskeletal, Forensic and Gastrointestinal. It is also better than \texttt{RadPhi2-Instruct} on all the systems. It is interesting to note that \texttt{Phi-3-mini-4k-instruct} has the highest recall value pointing to its reasonably good radiology knowledge and has comparable metrics to \texttt{GPT-4} on many systems as seen from the stratified metrics.

An example prediction from \texttt{RadPhi3-Instruct}, \texttt{Phi-3-mini-4k-instruct} and \texttt{GPT-4} models is available on Table \ref{tab:example_radqa}. We see that  \texttt{RadPhi3-Instruct} is able to give clear, concise and factually correct answer to the question.
\subsection{Impression Prediction}
The results of the evaluation of the impression prediction task from the findings of a radiology report on the test set of Mimic-CXR is available in Table \ref{tab:impression_prediction_metrics}. Both \texttt{RadPhi-2} and \texttt{RadPhi-3} is better than Impression GPT \citep{Ma_2024} and establishes SOTA performance. 
We note that we pre-processed the impressions with \texttt{GPT-4} to remove noise using the prompt we defined for the cleanup radiology text task in Table \ref{tab:prompt_design}.

The metrics of the impression prediction task on the CheXpert Plus validation dataset is available in Table \ref{tab:impression_prediction_metrics_chexpert}. 
An example prediction for the Impression prediction task is available in Table \ref{tab:example_impression_prediction} in the Appendix section.

\subsection{Abnormality and Support Devices Prediction}

The performance metrics  of the abnormality and support devices label prediction from the radiology reports is available in Table \ref{tab:label_prediction_metrics}. We see that \texttt{RadPhi-3} has comparable metrics to \texttt{RadPhi-2} in both the abnormality and support devices label prediction tasks. 
Example predictions for the   tubes, lines and devices prediction and abnormality labels prediction is available in Table \ref{tab:example_tubesnlines_label} and Table \ref{tab:example_abnormality_label} respectively.

\subsection{QA Comprehension, Cleanup Radiology Text and Findings/Impression Extraction Tasks}
The performance metrics of QA Comprehension and other utility tasks like noise removal from a radiology text and extracting a clean set of findings/impression from a radiology report is available in  Table \ref{tab:radinstruct_metrics}. We see that \texttt{RadPhi-3} has comparable metrics to \texttt{RadPhi-2} in the findings/impression extraction and cleanup radiology text tasks and \texttt{RadPhi-2} is slightly better on the QA comprehension task. We also note that the general domain \texttt{Phi3-mini-4k-instruct} model has lower performance on these tasks, particularly on the QA comprehension task. We also note the Align Score metric is high for these tasks indicating the factual consistency of the outputs to the inputs.

Example results for the QA comprehension task is available in Table \ref{tab:example_qacomprehension}, extract findings/impression task is available in Table \ref{tab:example_extraction} and cleanup radiology text task is available in Table \ref{tab:example_cleanup_radiology}.

\subsection{Natural Language Inference}

The performance metrics of Natural Language Inference on the MEDNLI and RADNLI datasets is available in Table \ref{tab:radnli_metrics}. 
\texttt{RadPhi-2} has poor performance on semantic tasks as it went through minimal general domain instruction tuning.  \texttt{RadPhi-3} shows significant improvement on this task over \texttt{RadPhi-2}. Sample results from the \texttt{RadPhi-3} is available in Table \ref{tab:example_RADNLI}. 
\subsection{Radiology Report Segmentation}
The performance of \texttt{RadPhi-3} on the CheXpert Plus \citep{chambon2024chexpertplusaugmentinglarge} validation dataset to divide the radiology report text into different sections (defined in Table \ref{tab:chexpert_dataset_stats}) is available in Table \ref{tab:chexpert_section_metrics}. We note that \texttt{RadPhi-3} performs very well on the radiology report section segmentation task, sample results are available in Table \ref{tab:example_chexpert_section_extraction}. The Align Score for the extraction of all the report sections is very high indicating the factual consistency of the outputs to the inputs.

\subsection{Temporal Change Summary}

The performance metrics of the Temporal Change Summary Task to summarise the changes in the medical conditions and support devices comparing the current radiology report and its prior is presented in Table \ref{tab:temporal_change_metrics}. As seen from the table, the best performing change category is `Stable' for both the medical conditions and support devices. The least performing categories for support devices are the `Changed' and `Recommendations' categories which reports the changes in the placement of devices and related recommendations. This is due to the fact that we have very less support for cases that report the changes in the placement of devices in the dataset. Sample model predictions for the temporal change summary task is available in Table \ref{tab:temporal_extraction_example}.

\subsection{Generalization Testing}

We also performed generalization testing of \texttt{Rad-Phi2-Instruct} and \texttt{Rad-Phi3-Instruct} models using some examples of tasks the models have not seen before. We have the following observations:

\paragraph{RadQA Dataset}

We performed a variant of the QA comprehension task using the RadQA \citep{Soni2022-vw} dataset. In this task, the model was asked to extract the part of the radiology report that can answer the question. The reports in this dataset pertain to various imaging modalities and anatomical systems. We found that \texttt{Rad-Phi3-Instruct} can generalize better than \texttt{Rad-Phi2-Instruct}. Example predictions from both models are available in Table \ref{tab:example_radqa_dataset}.

\paragraph{Sentence Similarity}

We used the MS-CXR-T \citep{Bannur2023-iy} dataset to test the generalization performance of \texttt{Rad-Phi3-Instruct} and \texttt{Rad-Phi2-Instruct} on a semantic task. The task is to assess whether two sentences are describing the same meaning (paraphrase) or different meanings (different) given a change summary. We found that \texttt{Rad-Phi2-Instruct} is unable to perform this task at all, while \texttt{Rad-Phi3-Instruct} and the base model \texttt{Phi3-mini-4k-instruct} were able to perform it. Example predictions are available in Table \ref{tab:example1_temporal_summary_mscxr} and Table \ref{tab:example2_temporal_summary_mscxr}.

\paragraph{Tubes and Lines Placement}

In this task, the models were asked to describe the placement of devices mentioned in the report. Although this task was new to the models, both \texttt{Rad-Phi2-Instruct} and \texttt{Rad-Phi3-Instruct} were able to perform it successfully, while the base model \texttt{Phi3-mini-4k-instruct} was not able to perform this task. Example predictions are available in Table \ref{tab:example_extraction_of_tubesandlines}.

\subsection{NLG Benchmarks}

\subsubsection{RaLEs Benchmark}
RaLEs is a benchmark group of datasets for natural language understanding and generation in radiology. We focus only on the natural language generation task from this benchmark to predict the Impressions section of a radiology report given the Findings, using the MEDIQA 2021 \citep{ben-abacha-etal-2021-overview} and BioNLP 2023 \citep{delbrouck-etal-2023-overview} datasets. BioNLP 2023  dataset contains computed tomography reports and magnetic resonance imaging examinations of head, chest, abdomen, spine and sinuses. This dataset is good to benchmark the impression prediction task for regions other than Chest and modalities other than X-rays. MEDIQA 2021 contains radiology reports of only Chest X-rays from the Mimic-CXR dataset. The statistics of the datasets in the RaLEs benchmark is available in Table \ref{tab:downstream_metrics}.

\paragraph{NLG Score}
We compute the NLG (Natural Language Generation) score for the RaLEs benchmark \citep{chaves2023RaLEs}. NLG score is the average of the ROUGE-L and RadGraph metrics across both the datasets for comparison with the leader board. 

We note that \texttt{RadPhi-3} achieves SOTA performance on the RaLEs benchmark surpassing the existing SOTA results by a large margin in both the clinical and lexical metrics for the BIONLP-2023 dataset. We also note that as \texttt{RadPhi-3} was trained on paraphrased versions of impressions of the MIMIC-CXR dataset, we preprocess the impression sections of the MEDIQA-2021 dataset which is also MIMIC-CXR based using the cleanup radiology text prompt defined in Table \ref{tab:prompt_design}. The metrics are available in Table \ref{tab:RaLEs_metrics}. 

\subsubsection{Temporal Radiology Report Summary}
We leverage the Radgraph2 dataset \citep{pmlr-v219-khanna23a} annotations to extract the lines from the radiology report that mentions a change about a medical condition or a support device.   The metrics are available in Table \ref{tab:radgraph2_metrics}. As seen from the result, the model is able to extract the change mentions of medical observations and support devices reasonably well for all categories except for the resolved category of medical observations which has very less support as seen from the dataset statistics of the RadGraph2 dataset in Table \ref{tab:radgraph2_stats}.  Sample predictions of this task is available in Table \ref{tab:example_temporal_extraction}.

\subsubsection{Padchest Benchmark}
We use the Padchest \citep{bustos2019padchestlargechestxray} dataset annotations which contains radiology reports in the Spanish language as a downstream task of \texttt{RadPhi-3} to predict the abnormality labels associated with the Spanish reports. We use a prompt in the Spanish language defined in Table \ref{tab:prompt_design} for this task. The metrics in Table \ref{tab:padchest_metrics} shows that \texttt{RadPhi-3}'s performance on this task is very good. A sample prediction on Spanish reports is available in Table \ref{tab:example_padchest}.

\section{Discussion}
\label{sec:discussion}

In this work, we presented \texttt{RadPhi-3}, an SLM  with 3.8B parameters finetuned from \texttt{Phi-3-mini-4k-instruct} for radiology specific tasks. The instruction tuning involved knowledge intensive tasks like radiology question answering and also ten different tasks related to radiology reports. Unlike \texttt{RadPhi-2}, which was finetuned from \texttt{Phi-2} and was  preceded with a general domain instruction tuning and used a special token based input format, \texttt{RadPhi-3} was directly finetuned from \texttt{Phi-3-mini-4k-instruct} using the default instruction format. \texttt{RadPhi-3} shows significant improvement over \texttt{RadPhi-2} on radiology question answering for all the systems and natural language inference tasks while being on-par on rest of the tasks.

SLMs like \texttt{RadPhi-3} can be easily trained for more nuanced tasks in Radiology domain where general domain LLMs falls short unless prompted with carefully curated dynamic examples \citep{Ma_2024}. Example of such tasks include long tail label prediction (Padchest radiology reports is mapped to 347 abnormality categories), impression prediction and radiology question answering which requires deep knowledge of the domain. This was also evident from the the stratified metrics of the radiology question answering task available in Table \ref{tab:metrics_radiopaedia_stratified} where we note that GPT-4 falls short in radiology knowledge when compared to \texttt{RadPhi-3} on systems like Hepatobiliary, Urogenital, Breast, Musculoskeletal, Forensic and Gastrointestinal. 

We presented two new tasks in this work, Radiology Report Segmentation to extract different sections from a radiology report and the Temporal Change Summary task to summarize the changes of medical observations and support devices from a radiology report and its prior report. We also bench marked \texttt{RadPhi-3}'s performance on ten different tasks like Impression Prediction, Abnormality Label Prediction, Tubes, Lines and Device Label Prediction, Temporal Summary Extraction, QA comprehension, Findings/Impression Extraction, Cleanup Radiology Text, Natural Language Inference, Radiology Report Segmentation and Abnormality Label Prediction from Spanish reports.

\texttt{RadPhi-3} achieves SOTA performance on the RaLEs benchmark for generative tasks with significant improvement for the BioNLP 2023 dataset which contains radiology reports from systems other than Chest and modalities other than X-rays showing that the model is able to generalize well and can be extended to other report types.
\paragraph{Limitations and Future Work} The current work did not explore the advantages of a radiology instruction-tuned SLM in a multimodal setting especially for finetuning in a limited dataset setting and low rank adaptation settings where we anticipate it would add more value. The future work can explore in this direction. \texttt{RadPhi-3} was also instruction tuned only using radiology reports of Chest X-rays, future work can extend this across systems, modalities, languages and additional tasks.



\acks

{
We thank the Radiopaedia team for their data and resources, which have advanced our radiology research. Special thanks to Dr. Kondaru Vishnu Vardhan Reddy from Manipal Hospitals Radiology Group, Bengaluru, and Dr.Sudhir Kumar Yadav from King George's Medical University, Lucknow, for evaluating our datasets and model predictions and providing valuable feedback.
}

\bibliography{jmlr-sample}

\begin{thebibliography}{28}
\providecommand{\natexlab}[1]{#1}
\providecommand{\url}[1]{\texttt{#1}}
\expandafter\ifx\csname urlstyle\endcsname\relax
  \providecommand{\doi}[1]{doi: #1}\else
  \providecommand{\doi}{doi: \begingroup \urlstyle{rm}\Url}\fi

\bibitem[Abdin et~al.(2024)Abdin, Aneja, Awadalla, Awadallah, Awan, Bach, Bahree, Bakhtiari, Bao, Behl, Benhaim, Bilenko, Bjorck, Bubeck, Cai, Cai, Chaudhary, Chen, Chen, Chen, Chen, Chen, Cheng, Chopra, Dai, Dixon, Eldan, Fragoso, Gao, Gao, Gao, Garg, Giorno, Goswami, Gunasekar, Haider, Hao, Hewett, Hu, Huynh, Iter, Jacobs, Javaheripi, Jin, Karampatziakis, Kauffmann, Khademi, Kim, Kim, Kurilenko, Lee, Lee, Li, Li, Liang, Liden, Lin, Lin, Liu, Liu, Liu, Liu, Liu, Luo, Madan, Mahmoudzadeh, Majercak, Mazzola, Mendes, Mitra, Modi, Nguyen, Norick, Patra, Perez-Becker, Portet, Pryzant, Qin, Radmilac, Ren, de~Rosa, Rosset, Roy, Ruwase, Saarikivi, Saied, Salim, Santacroce, Shah, Shang, Sharma, Shen, Shukla, Song, Tanaka, Tupini, Vaddamanu, Wang, Wang, Wang, Wang, Wang, Wang, Ward, Wen, Witte, Wu, Wu, Wyatt, Xiao, Xu, Xu, Xu, Xue, Yadav, Yang, Yang, Yang, Yang, Yu, Yuan, Zhang, Zhang, Zhang, Zhang, Zhang, Zhang, Zhang, and Zhou]{abdin2024phi3technicalreporthighly}
Marah Abdin, Jyoti Aneja, Hany Awadalla, Ahmed Awadallah, Ammar~Ahmad Awan, Nguyen Bach, Amit Bahree, Arash Bakhtiari, Jianmin Bao, Harkirat Behl, Alon Benhaim, Misha Bilenko, Johan Bjorck, Sébastien Bubeck, Martin Cai, Qin Cai, Vishrav Chaudhary, Dong Chen, Dongdong Chen, Weizhu Chen, Yen-Chun Chen, Yi-Ling Chen, Hao Cheng, Parul Chopra, Xiyang Dai, Matthew Dixon, Ronen Eldan, Victor Fragoso, Jianfeng Gao, Mei Gao, Min Gao, Amit Garg, Allie~Del Giorno, Abhishek Goswami, Suriya Gunasekar, Emman Haider, Junheng Hao, Russell~J. Hewett, Wenxiang Hu, Jamie Huynh, Dan Iter, Sam~Ade Jacobs, Mojan Javaheripi, Xin Jin, Nikos Karampatziakis, Piero Kauffmann, Mahoud Khademi, Dongwoo Kim, Young~Jin Kim, Lev Kurilenko, James~R. Lee, Yin~Tat Lee, Yuanzhi Li, Yunsheng Li, Chen Liang, Lars Liden, Xihui Lin, Zeqi Lin, Ce~Liu, Liyuan Liu, Mengchen Liu, Weishung Liu, Xiaodong Liu, Chong Luo, Piyush Madan, Ali Mahmoudzadeh, David Majercak, Matt Mazzola, Caio César~Teodoro Mendes, Arindam Mitra, Hardik Modi, Anh Nguyen,
  Brandon Norick, Barun Patra, Daniel Perez-Becker, Thomas Portet, Reid Pryzant, Heyang Qin, Marko Radmilac, Liliang Ren, Gustavo de~Rosa, Corby Rosset, Sambudha Roy, Olatunji Ruwase, Olli Saarikivi, Amin Saied, Adil Salim, Michael Santacroce, Shital Shah, Ning Shang, Hiteshi Sharma, Yelong Shen, Swadheen Shukla, Xia Song, Masahiro Tanaka, Andrea Tupini, Praneetha Vaddamanu, Chunyu Wang, Guanhua Wang, Lijuan Wang, Shuohang Wang, Xin Wang, Yu~Wang, Rachel Ward, Wen Wen, Philipp Witte, Haiping Wu, Xiaoxia Wu, Michael Wyatt, Bin Xiao, Can Xu, Jiahang Xu, Weijian Xu, Jilong Xue, Sonali Yadav, Fan Yang, Jianwei Yang, Yifan Yang, Ziyi Yang, Donghan Yu, Lu~Yuan, Chenruidong Zhang, Cyril Zhang, Jianwen Zhang, Li~Lyna Zhang, Yi~Zhang, Yue Zhang, Yunan Zhang, and Xiren Zhou.
\newblock Phi-3 technical report: A highly capable language model locally on your phone, 2024.
\newblock URL \url{https://arxiv.org/abs/2404.14219}.

\bibitem[Bannur et~al.(2023)Bannur, Hyland, Liu, P{\'e}rez-Garc{\'\i}a, Ilse, Coelho~de Castro, Boecking, Sharma, Bouzid, Schwaighofer, Wetscherek, Richardson, Naumann, Alvarez~Valle, and Oktay]{Bannur2023-iy}
Shruthi Bannur, Stephanie Hyland, Qianchu Liu, Fernando P{\'e}rez-Garc{\'\i}a, Max Ilse, Daniel Coelho~de Castro, Benedikt Boecking, Harshita Sharma, Kenza Bouzid, Anton Schwaighofer, Maria~Teodora Wetscherek, Hannah Richardson, Tristan Naumann, Javier Alvarez~Valle, and Ozan Oktay.
\newblock {MS-CXR-T}: Learning to exploit temporal structure for biomedical vision-language processing, 2023.

\bibitem[Ben~Abacha et~al.(2021)Ben~Abacha, Mrabet, Zhang, Shivade, Langlotz, and Demner-Fushman]{ben-abacha-etal-2021-overview}
Asma Ben~Abacha, Yassine Mrabet, Yuhao Zhang, Chaitanya Shivade, Curtis Langlotz, and Dina Demner-Fushman.
\newblock Overview of the {MEDIQA} 2021 shared task on summarization in the medical domain.
\newblock In Dina Demner-Fushman, Kevin~Bretonnel Cohen, Sophia Ananiadou, and Junichi Tsujii, editors, \emph{Proceedings of the 20th Workshop on Biomedical Language Processing}, pages 74--85, Online, June 2021. Association for Computational Linguistics.
\newblock \doi{10.18653/v1/2021.bionlp-1.8}.
\newblock URL \url{https://aclanthology.org/2021.bionlp-1.8}.

\bibitem[Bustos et~al.(2019)Bustos, Pertusa, Salinas, and de~la Iglesia-Vayá]{bustos2019padchestlargechestxray}
Aurelia Bustos, Antonio Pertusa, Jose-Maria Salinas, and Maria de~la Iglesia-Vayá.
\newblock Padchest: A large chest x-ray image dataset with multi-label annotated reports, 2019.
\newblock URL \url{https://arxiv.org/abs/1901.07441}.

\bibitem[Chambon et~al.(2024)Chambon, Delbrouck, Sounack, Huang, Chen, Varma, Truong, Chuong, and Langlotz]{chambon2024chexpertplusaugmentinglarge}
Pierre Chambon, Jean-Benoit Delbrouck, Thomas Sounack, Shih-Cheng Huang, Zhihong Chen, Maya Varma, Steven~QH Truong, Chu~The Chuong, and Curtis~P. Langlotz.
\newblock Chexpert plus: Augmenting a large chest x-ray dataset with text radiology reports, patient demographics and additional image formats, 2024.
\newblock URL \url{https://arxiv.org/abs/2405.19538}.

\bibitem[Chaves et~al.(2023)Chaves, Bhaskhar, Attias, Delbrouck, Rubin, Loening, Langlotz, and Chaudhari]{chaves2023RaLEs}
Juan Manuel~Zambrano Chaves, Nandita Bhaskhar, Maayane Attias, Jean-Benoit Delbrouck, Daniel Rubin, Andreas~Markus Loening, Curtis Langlotz, and Akshay~S Chaudhari.
\newblock Ra{LE}s: a benchmark for radiology language evaluations.
\newblock In \emph{Thirty-seventh Conference on Neural Information Processing Systems Datasets and Benchmarks Track}, 2023.
\newblock URL \url{https://openreview.net/forum?id=PWLGrvoqiR}.

\bibitem[Dejl et~al.(2024)Dejl, Khanna, Pile, Yoon, Truong, Duong, Saenz, and Rajpurkar]{Dejl2024-mt}
Adam Dejl, Sameer Khanna, Patricia~Therese Pile, Kibo Yoon, Steven Q~H Truong, Hanh Duong, Agustina Saenz, and Pranav Rajpurkar.
\newblock {RadGraph2}: Tracking findings over time in radiology reports, 2024.

\bibitem[Delbrouck et~al.(2023)Delbrouck, Varma, Chambon, and Langlotz]{delbrouck-etal-2023-overview}
Jean-Benoit Delbrouck, Maya Varma, Pierre Chambon, and Curtis Langlotz.
\newblock Overview of the {R}ad{S}um23 shared task on multi-modal and multi-anatomical radiology report summarization.
\newblock In Dina Demner-fushman, Sophia Ananiadou, and Kevin Cohen, editors, \emph{The 22nd Workshop on Biomedical Natural Language Processing and BioNLP Shared Tasks}, pages 478--482, Toronto, Canada, July 2023. Association for Computational Linguistics.
\newblock \doi{10.18653/v1/2023.bionlp-1.45}.
\newblock URL \url{https://aclanthology.org/2023.bionlp-1.45}.

\bibitem[Hu et~al.(2023)Hu, Gu, An, Zhang, Liu, Kobayashi, Harada, Summers, and Zhu]{10.1145/3580305.3599819}
Xinyue Hu, Lin Gu, Qiyuan An, Mengliang Zhang, Liangchen Liu, Kazuma Kobayashi, Tatsuya Harada, Ronald~M. Summers, and Yingying Zhu.
\newblock Expert knowledge-aware image difference graph representation learning for difference-aware medical visual question answering.
\newblock In \emph{Proceedings of the 29th ACM SIGKDD Conference on Knowledge Discovery and Data Mining}, KDD '23, page 4156–4165, New York, NY, USA, 2023. Association for Computing Machinery.
\newblock ISBN 9798400701030.
\newblock \doi{10.1145/3580305.3599819}.
\newblock URL \url{https://doi.org/10.1145/3580305.3599819}.

\bibitem[Jain et~al.(2021)Jain, Agrawal, Saporta, Truong, Duong, Bui, Chambon, Zhang, Lungren, Ng, Langlotz, and Rajpurkar]{jain2021radgraphextractingclinicalentities}
Saahil Jain, Ashwin Agrawal, Adriel Saporta, Steven~QH Truong, Du~Nguyen Duong, Tan Bui, Pierre Chambon, Yuhao Zhang, Matthew~P. Lungren, Andrew~Y. Ng, Curtis~P. Langlotz, and Pranav Rajpurkar.
\newblock Radgraph: Extracting clinical entities and relations from radiology reports, 2021.
\newblock URL \url{https://arxiv.org/abs/2106.14463}.

\bibitem[Johnson et~al.(2019)Johnson, Pollard, Berkowitz, Greenbaum, Lungren, Deng, Mark, and Horng]{Johnson2019}
Alistair E.~W. Johnson, Tom~J. Pollard, Seth~J. Berkowitz, Nathaniel~R. Greenbaum, Matthew~P. Lungren, Chih-ying Deng, Roger~G. Mark, and Steven Horng.
\newblock Mimic-cxr, a de-identified publicly available database of chest radiographs with free-text reports.
\newblock \emph{Scientific Data}, 6\penalty0 (1):\penalty0 317, Dec 2019.
\newblock ISSN 2052-4463.
\newblock \doi{10.1038/s41597-019-0322-0}.
\newblock URL \url{https://doi.org/10.1038/s41597-019-0322-0}.

\bibitem[Khanna et~al.(2023)Khanna, Dejl, Yoon, Truong, Duong, Saenz, and Rajpurkar]{pmlr-v219-khanna23a}
Sameer Khanna, Adam Dejl, Kibo Yoon, Steven~QH Truong, Hanh Duong, Agustina Saenz, and Pranav Rajpurkar.
\newblock Radgraph2: Modeling disease progression in radiology reports via hierarchical information extraction.
\newblock In Kaivalya Deshpande, Madalina Fiterau, Shalmali Joshi, Zachary Lipton, Rajesh Ranganath, Iñigo Urteaga, and Serene Yeung, editors, \emph{Proceedings of the 8th Machine Learning for Healthcare Conference}, volume 219 of \emph{Proceedings of Machine Learning Research}, pages 381--402. PMLR, 11--12 Aug 2023.
\newblock URL \url{https://proceedings.mlr.press/v219/khanna23a.html}.

\bibitem[Li et~al.(2023)Li, Wong, Zhang, Usuyama, Liu, Yang, Naumann, Poon, and Gao]{li2023llavamedtraininglargelanguageandvision}
Chunyuan Li, Cliff Wong, Sheng Zhang, Naoto Usuyama, Haotian Liu, Jianwei Yang, Tristan Naumann, Hoifung Poon, and Jianfeng Gao.
\newblock Llava-med: Training a large language-and-vision assistant for biomedicine in one day, 2023.
\newblock URL \url{https://arxiv.org/abs/2306.00890}.

\bibitem[Lin(2004)]{lin-2004-rouge}
Chin-Yew Lin.
\newblock {ROUGE}: A package for automatic evaluation of summaries.
\newblock In \emph{Text Summarization Branches Out}, pages 74--81, Barcelona, Spain, July 2004. Association for Computational Linguistics.
\newblock URL \url{https://aclanthology.org/W04-1013}.

\bibitem[Liu et~al.(2023{\natexlab{a}})Liu, Hyland, Bannur, Bouzid, Castro, Wetscherek, Tinn, Sharma, P{\'e}rez-Garc{\'\i}a, Schwaighofer, Rajpurkar, Khanna, Poon, Usuyama, Thieme, Nori, Lungren, Oktay, and Alvarez-Valle]{liu-etal-2023-exploring-boundaries}
Qianchu Liu, Stephanie Hyland, Shruthi Bannur, Kenza Bouzid, Daniel Castro, Maria Wetscherek, Robert Tinn, Harshita Sharma, Fernando P{\'e}rez-Garc{\'\i}a, Anton Schwaighofer, Pranav Rajpurkar, Sameer Khanna, Hoifung Poon, Naoto Usuyama, Anja Thieme, Aditya Nori, Matthew Lungren, Ozan Oktay, and Javier Alvarez-Valle.
\newblock Exploring the boundaries of {GPT}-4 in radiology.
\newblock In Houda Bouamor, Juan Pino, and Kalika Bali, editors, \emph{Proceedings of the 2023 Conference on Empirical Methods in Natural Language Processing}, pages 14414--14445, Singapore, December 2023{\natexlab{a}}. Association for Computational Linguistics.
\newblock \doi{10.18653/v1/2023.emnlp-main.891}.
\newblock URL \url{https://aclanthology.org/2023.emnlp-main.891}.

\bibitem[Liu et~al.(2023{\natexlab{b}})Liu, Li, Shu, Zhong, Yang, Ju, Wu, Ma, Luo, Chen, Kim, Hu, Dai, Zhao, Zhu, Liu, Liu, Shen, Liu, Li, and Li]{liu2023radiologyllama2bestinclasslargelanguage}
Zhengliang Liu, Yiwei Li, Peng Shu, Aoxiao Zhong, Longtao Yang, Chao Ju, Zihao Wu, Chong Ma, Jie Luo, Cheng Chen, Sekeun Kim, Jiang Hu, Haixing Dai, Lin Zhao, Dajiang Zhu, Jun Liu, Wei Liu, Dinggang Shen, Tianming Liu, Quanzheng Li, and Xiang Li.
\newblock Radiology-llama2: Best-in-class large language model for radiology, 2023{\natexlab{b}}.
\newblock URL \url{https://arxiv.org/abs/2309.06419}.

\bibitem[Liu et~al.(2024)Liu, Zhong, Li, Yang, Ju, Wu, Ma, Shu, Chen, Kim, Dai, Zhao, Sun, Zhu, Liu, Liu, Shen, Li, Li, and Liu]{liu2024radiologygptlargelanguagemodel}
Zhengliang Liu, Aoxiao Zhong, Yiwei Li, Longtao Yang, Chao Ju, Zihao Wu, Chong Ma, Peng Shu, Cheng Chen, Sekeun Kim, Haixing Dai, Lin Zhao, Lichao Sun, Dajiang Zhu, Jun Liu, Wei Liu, Dinggang Shen, Xiang Li, Quanzheng Li, and Tianming Liu.
\newblock Radiology-gpt: A large language model for radiology, 2024.
\newblock URL \url{https://arxiv.org/abs/2306.08666}.

\bibitem[Ma et~al.(2024)Ma, Wu, Wang, Xu, Wei, Liu, Zeng, Jiang, Guo, Cai, Zhang, Zhang, Zhu, Shen, Liu, and Li]{Ma_2024}
Chong Ma, Zihao Wu, Jiaqi Wang, Shaochen Xu, Yaonai Wei, Zhengliang Liu, Fang Zeng, Xi~Jiang, Lei Guo, Xiaoyan Cai, Shu Zhang, Tuo Zhang, Dajiang Zhu, Dinggang Shen, Tianming Liu, and Xiang Li.
\newblock An iterative optimizing framework for radiology report summarization with chatgpt.
\newblock \emph{IEEE Transactions on Artificial Intelligence}, 5\penalty0 (8):\penalty0 4163–4175, August 2024.
\newblock ISSN 2691-4581.
\newblock \doi{10.1109/tai.2024.3364586}.
\newblock URL \url{http://dx.doi.org/10.1109/TAI.2024.3364586}.

\bibitem[Miura et~al.(2021)Miura, Zhang, Tsai, Langlotz, and Jurafsky]{Miura2021-bl}
Yasuhide Miura, Yuhao Zhang, Emily Tsai, Curtis Langlotz, and Dan Jurafsky.
\newblock {RadNLI}: A natural language inference dataset for the radiology domain, 2021.

\bibitem[Ouyang et~al.(2022)Ouyang, Wu, Jiang, Almeida, Wainwright, Mishkin, Zhang, Agarwal, Slama, Ray, Schulman, Hilton, Kelton, Miller, Simens, Askell, Welinder, Christiano, Leike, and Lowe]{ouyang2022traininglanguagemodelsfollow}
Long Ouyang, Jeff Wu, Xu~Jiang, Diogo Almeida, Carroll~L. Wainwright, Pamela Mishkin, Chong Zhang, Sandhini Agarwal, Katarina Slama, Alex Ray, John Schulman, Jacob Hilton, Fraser Kelton, Luke Miller, Maddie Simens, Amanda Askell, Peter Welinder, Paul Christiano, Jan Leike, and Ryan Lowe.
\newblock Training language models to follow instructions with human feedback, 2022.
\newblock URL \url{https://arxiv.org/abs/2203.02155}.

\bibitem[Papineni et~al.(2002)Papineni, Roukos, Ward, and Zhu]{papineni-etal-2002-bleu}
Kishore Papineni, Salim Roukos, Todd Ward, and Wei-Jing Zhu.
\newblock {B}leu: a method for automatic evaluation of machine translation.
\newblock In Pierre Isabelle, Eugene Charniak, and Dekang Lin, editors, \emph{Proceedings of the 40th Annual Meeting of the Association for Computational Linguistics}, pages 311--318, Philadelphia, Pennsylvania, USA, July 2002. Association for Computational Linguistics.
\newblock \doi{10.3115/1073083.1073135}.
\newblock URL \url{https://aclanthology.org/P02-1040}.

\bibitem[Ranjit et~al.(2024)Ranjit, Ganapathy, Srivastav, Oruganti, and Ganu]{Ranjit2024-pk}
Mercy Ranjit, Gopinath Ganapathy, Shaury Srivastav, Srujana Oruganti, and Tanuja Ganu.
\newblock {Rad-Phi2}: Instruction tuning phi2 for radiology.
\newblock \emph{Advances in Artificial Intelligence and Machine Learning}, 04\penalty0 (02):\penalty0 2302--2323, 2024.

\bibitem[Shivade(2017)]{Shivade2017-zu}
Chaitanya Shivade.
\newblock {MedNLI} --- a natural language inference dataset for the clinical domain, 2017.

\bibitem[Smit et~al.(2020)Smit, Jain, Rajpurkar, Pareek, Ng, and Lungren]{smit-etal-2020-combining}
Akshay Smit, Saahil Jain, Pranav Rajpurkar, Anuj Pareek, Andrew Ng, and Matthew Lungren.
\newblock Combining automatic labelers and expert annotations for accurate radiology report labeling using {BERT}.
\newblock In Bonnie Webber, Trevor Cohn, Yulan He, and Yang Liu, editors, \emph{Proceedings of the 2020 Conference on Empirical Methods in Natural Language Processing (EMNLP)}, pages 1500--1519, Online, November 2020. Association for Computational Linguistics.
\newblock \doi{10.18653/v1/2020.emnlp-main.117}.
\newblock URL \url{https://aclanthology.org/2020.emnlp-main.117}.

\bibitem[Soni and Roberts(2022)]{Soni2022-vw}
Sarvesh Soni and Kirk Roberts.
\newblock {RadQA}: A question answering dataset to improve comprehension of radiology reports, 2022.

\bibitem[Wu et~al.(2021)Wu, Agu, Lourentzou, Sharma, Paguio, Yao, Dee, Mitchell, Kashyap, Giovannini, Celi, Syeda-Mahmood, and Moradi]{Wu2021-sc}
Joy Wu, Nkechinyere Agu, Ismini Lourentzou, Arjun Sharma, Joseph Paguio, Jasper~Seth Yao, Edward~Christopher Dee, William Mitchell, Satyananda Kashyap, Andrea Giovannini, Leo~Anthony Celi, Tanveer Syeda-Mahmood, and Mehdi Moradi.
\newblock Chest {ImaGenome} dataset, 2021.

\bibitem[Zha et~al.(2023)Zha, Yang, Li, and Hu]{zha2023alignscoreevaluatingfactualconsistency}
Yuheng Zha, Yichi Yang, Ruichen Li, and Zhiting Hu.
\newblock Alignscore: Evaluating factual consistency with a unified alignment function, 2023.
\newblock URL \url{https://arxiv.org/abs/2305.16739}.

\bibitem[Zhang et~al.(2023)Zhang, Wang, Yang, Yu, Vu, and Lei]{zhang-etal-2023-long}
Ruohong Zhang, Yau-Shian Wang, Yiming Yang, Donghan Yu, Tom Vu, and Likun Lei.
\newblock Long-tailed extreme multi-label text classification by the retrieval of generated pseudo label descriptions.
\newblock In Andreas Vlachos and Isabelle Augenstein, editors, \emph{Findings of the Association for Computational Linguistics: EACL 2023}, pages 1092--1106, Dubrovnik, Croatia, May 2023. Association for Computational Linguistics.
\newblock \doi{10.18653/v1/2023.findings-eacl.81}.
\newblock URL \url{https://aclanthology.org/2023.findings-eacl.81}.

\end{thebibliography}

\appendix
\onecolumn
\section{Evaluation Benchmarks}
\label{apd:eval_config}

\begin{table*}[!ht]
\caption{Radiology Report Segmentation Metrics (CheXpert Plus sections)}
\centering
\footnotesize
\begin{tabular}{|lcccc|}
\hline
\textbf{Report Section} & \textbf{BLEU-4} & \textbf{RougeL} & \textbf{F1-Score} & \textbf{AlignScore} \\
\hline
Clinical History & 66.38 & 100.0 & 98.56 & 99.46 \\
Comparisons & 73.39 & 100.00 &  96.49 & 99.80 \\
Technique & 8.54 & 100.00 &  98.25 & 99.36 \\
Procedure Comments & 16.41 & 100.00 &  98.37 & 99.41 \\
Impression & 99.81 & 99.91 &  99.84 & 89.03 \\
Findings & 32.71 & 99.01 &  98.27 & 98.88 \\
End Of Impression & 4.78 & 98.72 &  97.57 & 99.31 \\
Summary & 83.03 & 100  & 99.61 & 99.69 \\
\bottomrule
\multicolumn{5}{c}{\texttt{RadPhi-3} performs very well on section extraction task.}
\end{tabular}
\label{tab:chexpert_section_metrics}
\end{table*}

\begin{table*}[!ht]
\centering
\footnotesize
    \caption{Temporal Change Summary Metrics (Mimic-CXR reports)}
    \begin{tabular}{|l p{80pt} c c c c|}
        \hline
        \textbf{Change Category} &\textbf{Labels}&\textbf{BLEU-4} & \textbf{RougeL} & \textbf{F1-Score} & \textbf{Support}\\
        \hline

        Medical conditions Change Summary&
        \begin{tabular}{l}New \\ Resolved \\  Stable \\Improved \\ Worsened \\ Negatives \end{tabular}
        &\begin{tabular}{l}20.13 \\10.44 \\ 41.24 \\ 8.30 \\ 8.18 \\ 43.03 \end{tabular}
        &\begin{tabular}{l} 
        27.74 \\ 17.62\\ 61.94 \\ 13.98 \\ 13.19 \\ 65.97\\
        \end{tabular}
        &\begin{tabular}{l} 
        30.10 \\ 18.79 \\ 67.73 \\ 14.75 \\ 14.45 \\ 70.61
        \end{tabular}
        &\begin{tabular}{l} 
        1243 \\ 840 \\ 2564 \\ 543 \\ 666 \\ 2374
        \end{tabular}\\
        \hline
        
Tubes and Lines Change Summary&
        \begin{tabular}{l}New \\ Removed \\  Unchanged \\ Changed \\ Recommendations \end{tabular}
        &\begin{tabular}{l}16.73 \\11.13 \\ 23.61 \\ 1.70 \\ 3.74  \end{tabular}
        &\begin{tabular}{l} 
        19.32 \\ 18.96\\ 29.94 \\ 1.99 \\ 4.20\\
        \end{tabular}
        &\begin{tabular}{l} 
        20.10 \\ 20.10 \\ 31.21 \\ 2.09 \\ 4.40
        \end{tabular}
        &\begin{tabular}{l} 
        754 \\ 772 \\ 1177 \\ 132 \\ 171
        \end{tabular}\\
        \hline
        
            \multicolumn{6}{p{400pt}}{Performance of \texttt{RadPhi-3} on Temporal Change Summary task. Low performance in Changed and Recommendations category can be attributed to less support.}
    \end{tabular}   
    \label{tab:temporal_change_metrics}
\end{table*}

\begin{table*}[!ht]
\centering
\footnotesize
    \caption{RaLEs Benchmark - Impression Prediction}
    \begin{tabular}{ |l c c c|}
        \hline
        \textbf{Dataset} &\textbf{Metrics}& \textbf{Best Reported} &\textbf{RadPhi-3}\\
        \hline

        MEDIQA-2021 &
        \begin{tabular}{l}Rouge 2\\RougeL\\ Chexbert \\RG\end{tabular}
        &\begin{tabular}{l} 0.250 \\ 0.388  \\ \textbf{0.725}\\ \textbf{0.406} \end{tabular}
        &\begin{tabular}{l} \textbf{0.271} \\ \textbf{0.409} \\ 0.697\\ 0.353\end{tabular}
        \\
        \hline
        BIONLP-2023 &
        \begin{tabular}{l}Rouge 2\\RougeL\\ Chexbert \\RG\end{tabular}
        &
        \begin{tabular}{l}0.189 \\ 0.303\\ 0.506 \\ 0.283
        \end{tabular}&
        \begin{tabular}{l} \textbf{0.204} \\ \textbf{0.310} \\ \textbf{0.564} \\ \textbf{0.325}\end{tabular}
        \\
        \hline
        \textbf{NLG Score} & & \begin{tabular}{l}0.345\end{tabular} & \begin{tabular}{l} \textbf{0.349} \end{tabular}
        \\
        \hline
    \end{tabular}
    \label{tab:RaLEs_metrics}
\end{table*}

\begin{table*}[!ht]
\centering
\footnotesize
    \caption{Padchest Label Metrics (Spanish Reports)}
    \begin{tabular}{|lc|}
    \hline
        \textbf{Metric} & \textbf{Score}\\
        \hline
        F1-score & 97.64 {\scriptsize \textcolor{gray}{[97.51, 97.71]}}\\
        Precision & 97.79 {\scriptsize \textcolor{gray}{[97.67, 97.88]}}\\
        Recall & 97.9 {\scriptsize \textcolor{gray}{[97.78, 97.96]}}\\
        \hline
    \end{tabular}   
    \label{tab:padchest_metrics}
\end{table*}

\begin{table*}[!ht]
\centering
\footnotesize
    \caption{Temporal Change Summary Metrics - RadGraph2}
    \begin{tabular}{|l c c c c c |}
        \hline
        \textbf{Change Category} &\textbf{Progression}&\textbf{BLEU-4} & \textbf{RougeL} &\textbf{F1-Score}
        & \textbf{Support}\\
        \hline

        Medical conditions and devices &
        \begin{tabular}{l}Stable \end{tabular}
        &
        \begin{tabular}{l}86.30 \end{tabular} 
        &
        \begin{tabular}{l}81.60  \end{tabular} 
        &
        \begin{tabular}{l}93.99\end{tabular} 
        &
        77
        \\
        \hline
        Medical conditions &
        \begin{tabular}{l}Improved\\ Worsened \\ New \\ Resolved\end{tabular}
        &
        \begin{tabular}{l}83.61 \\ 81.18 \\ 70.45 \\ 36.46 \end{tabular} 
        &
        \begin{tabular}{l}81.85 \\79.13 \\ 72.13\\ 43.08\end{tabular} 
        &
        \begin{tabular}{l}89.09 \\86.61 \\ 77.99\\ 46.71\end{tabular} 
        &
        \begin{tabular}{l} 13 \\ 21 \\ 8 \\ 1 \end{tabular}
        \\
        \hline
        
        Medical devices & 
        \begin{tabular}{l}Removed \\ New \\ Advanced \end{tabular}
        &
        \begin{tabular}{l}87.56 \\  43.13 \\ 82.76\end{tabular} 
        &
        \begin{tabular}{l} 90.94\\ 52.13 \\ 85.19 \end{tabular} 
        &
        \begin{tabular}{l}  96.43\\ 55.17\\   88.25\end{tabular} 
        &
        \begin{tabular}{l} 7 \\ 5 \\ 2 \end{tabular}
        \\
        \bottomrule
                \multicolumn{6}{p{400pt}}{Performance of \texttt{RadPhi-3} on Temporal Change Summary task. Low performance on Resolved category due to very less support.}
    \end{tabular}   
    
    \label{tab:radgraph2_metrics}
\end{table*}

\section{Dataset Statistics}\label{apd:first}

\begin{table*}[!ht]
\centering
\footnotesize
\caption{Radiology Question Answering Dataset Statistics (Radiopaedia.org)}
\begin{tabular}{|l r r r|}
\hline
\textbf{Systems} & \textbf{Article Counts} & \textbf{Summary Counts} & \textbf{QA Pair Counts}\\
\hline
Chest&1710&31&9695\\
Cardiac&767&4&5188\\
Central Nervous System&2817&29&16557\\
Urogenital&703&2&4171\\
Oncology&391&0&2782\\
Breast&356&1&1929\\
Musculoskeletal&3758&46&24026\\
Not Specified&1579&0&10880\\
Hepatobiliary&469&1&2956\\
Vascular&420&3&2304\\
Gastrointestinal&1210&30&7342\\
Obstetrics&568&0&3157\\
Interventional&149&0&1003\\
Trauma&85&0&483\\
Spine&94&0&490\\
Forensic&17&1&105\\
\hline
\textbf{Total}&15076&148&93068\\
\hline
\end{tabular}
\label{tab:radiopaedia_stats}
\end{table*}

\begin{table*}[!ht]
\centering
\footnotesize
\caption{Radiology Report Task Dataset Statistics}
\begin{tabular}{|lccc|}
\hline
\textbf{Task} & \textbf{Train} & \textbf{Test} & \textbf{Validation} \\
\hline
Impression Prediction & 208,876 & 2,523 & 1,647 \\
Cleanup Radiology Text & 50,000 & 9,337 & 5,549\\
Abnormality Labels & 221,035 & 3,403 & 1,959 \\
QA Comprehension & 467,057 & 9,179 & 3,878 \\
Extract Findings & 50,000 & 3,844 & 2,103 \\
Tubes, Lines and Devices Labels & 94,915 & 3,403 & 1,959 \\
Extract Impression & 50,000 & 3,283 & 2,295 \\
RadNLI, MEDNLI & 11,712 & 1,902 & 1,395 \\
Radiology Report Segmentation & 282,592 & - & 308 \\
Impression Prediction (Chexpert Plus) & 59,364 & - & 74\\
Temporal Change Summary &151667 & 2,817 & 1,259\\
\bottomrule
\end{tabular}

\label{tab:radinstruct_dataset_stats}
\end{table*}

\begin{table*}[!ht]
\centering
\footnotesize
\caption{Benchmark Dataset Statistics}
\begin{tabular}{|lccc|}
\hline
\textbf{Dataset} & \textbf{Train} & \textbf{Test} & \textbf{Validation} \\
\hline
BioNLP-2023 (RaLEs) & 59,320 & 6,526 & 7,413 \\
MEDIQA-2021 (RaLEs) & 91,544 & 600 & 2,000\\
Temporal Change Summary (Radgraph-2) & 227,643 & 150 & 75 \\
Padchest Label Prediction & 137,221 & 7,872 & 15,665\\
\bottomrule
\end{tabular}

\label{tab:downstream_metrics}
\end{table*}

\begin{table*}
\centering
\footnotesize
    \caption{Temporal Change Summary Statistics - RadGraph2}
    \begin{tabular}{|l c c c c|}
        \hline
        \textbf{Change Summary} &\textbf{Progression}&\textbf{Train} & \textbf{Test} &\textbf{Validation}\\
        \hline

        Medical conditions and devices &
        \begin{tabular}{l}Stable \end{tabular}
        &
        \begin{tabular}{l}381 \end{tabular} 
        &
        \begin{tabular}{l}77  \end{tabular} 
        &
        \begin{tabular}{l}47 \end{tabular}
        \\
        \hline
        Medical conditions &
        \begin{tabular}{l}Improved\\ Worsened \\ New \\ Resolved\end{tabular}
        &
        \begin{tabular}{l}86 \\ 127 \\ 49 \\ 16 \end{tabular} 
        &
        \begin{tabular}{l}13 \\21 \\ 8\\ 1\end{tabular} 
        &
        \begin{tabular}{l}15 \\17 \\ 3\\ 2\end{tabular} 
        \\
        \hline
        
        Medical devices & 
        \begin{tabular}{l}Removed \\ New \\ Advanced \end{tabular}
        &
        \begin{tabular}{l}33 \\  17 \\ 13\end{tabular} 
        &
        \begin{tabular}{l} 7\\5 \\ 2 \end{tabular} 
        &
        \begin{tabular}{l}  6\\ 3\\   1\end{tabular} 
        \\
        \bottomrule
    \end{tabular}   
    \label{tab:radgraph2_stats}
\end{table*}

\begin{table*}[!ht]
\centering
\footnotesize
    \caption{CheXpert Plus Section Statistics}
    \begin{tabular}{|lcc|}
        \hline
        \textbf{Section} & \textbf{Train} & \textbf{Validation} \\
        \hline
        Comparison & 212,845 & 228 \\
        Impression & 223,083 & 234 \\
        Summary & 189,455 & 193 \\
        Clinical History & 176,871 & 175 \\
        Finding & 59,395 & 74 \\
        Procedure Comments & 25,637 & 32 \\
        Technique & 10,719 & 13 \\
        End of Impression & 7,385 & 7 \\
        \hline
        Total Records & 282,592 & 308 \\
        \bottomrule
    \end{tabular}
    \label{tab:chexpert_dataset_stats}
\end{table*}

\begin{table*}[!ht]
\centering
\footnotesize
    \caption{Temporal Change Summary - Mimic CXR}
    \begin{tabular}{|l l c c c|}
        \hline
        \textbf{Summary Category} &\textbf{Labels}&\textbf{Train} & \textbf{Test} &\textbf{Validation}\\
        \hline

        Diseases Change\_Summary&
        \begin{tabular}{l}Stable \\ Resolved \\  Negatives \\Worsened \\ New \\ Improved \end{tabular}
        &\begin{tabular}{l}303,689 \\68,939 \\ 222,780 \\ 40,827 \\ 116,147 \\ 34,318 \end{tabular}
        &\begin{tabular}{l} 
        6,150 \\ 1,323\\ 4,102 \\ 864 \\ 2,409 \\ 683\\
        \end{tabular}
        &\begin{tabular}{l} 
        2,604 \\ 589 \\ 1,848 \\ 340 \\ 970 \\ 284
        \end{tabular}\\
        \hline
        Tube\_Lines Change\_Summary&
        \begin{tabular}{l}Removed \\ Unchanged \\  New \\Recommendations \\ Changed \end{tabular}
        &\begin{tabular}{l} 51,738 \\69,263 \\ 49,940 \\ 10,110 \\ 9,637 \end{tabular}
        &\begin{tabular}{l} 
        1,029 \\ 1,554\\ 1,077 \\ 183 \\ 141\\
        \end{tabular}
        &\begin{tabular}{l} 
        432 \\ 642 \\ 441 \\ 91 \\ 77
        \end{tabular}\\
        \hline
    \end{tabular}   
    \label{tab:temporal_dataset_stats}
\end{table*}

\clearpage
\section{Stratified Metrics}\label{apd:second}

\begin{table*}[!ht]
\centering
\footnotesize
    \caption{System Wise Metrics for Radiology Question Answering on Radiopaedia Summary Articles.}
    \label{tab:metrics_radiopaedia_stratified}
    \resizebox{\textwidth}{!}{ 
    \begin{tabular}{|lrrrrr|}
        \hline
        \textbf{System} & \textbf{Metrics} & \textbf{Phi-3-mini-4k-instruct}& \textbf{RadPhi-2} & \textbf{GPT-4}& \textbf{RadPhi-3} \\
        \hline

        Chest
        &
        \begin{tabular}{l}
        F1-Score \\
        Precision\\
        Recall \\
        RougeL
        \end{tabular}
        &
        \begin{tabular}{l}
        35.08 {\scriptsize \textcolor{gray}{[34.24, 36.12]}}\\
        22.87 {\scriptsize \textcolor{gray}{[21.81, 23.47]}}\\
        \textbf{75.22 {\scriptsize \textcolor{gray}{[74.43, 76.54]}}}\\
        12.33 {\scriptsize \textcolor{gray}{[11.55, 12.69]}}
        \end{tabular}
        &
        \begin{tabular}{l}
        \textbf{39.64  {\scriptsize \textcolor{gray}{[37.68, 41.64]}}}\\
        42.76 {\scriptsize \textcolor{gray}{[40.56, 45.37]}}\\
        36.95 {\scriptsize \textcolor{gray}{[34.67, 39.66]}}\\
        26.83 {\scriptsize \textcolor{gray}{[25.33, 28.54]}}
        \end{tabular}
        &
        \begin{tabular}{l}
        36.73  {\scriptsize \textcolor{gray}{[35.39, 37.04]}}\\
        32.41 {\scriptsize \textcolor{gray}{[30.62, 33.62]}}\\
        42.37 {\scriptsize \textcolor{gray}{[40.13, 43.81]}}\\
        24.65 {\scriptsize \textcolor{gray}{[23.86, 25.54]}}
        \end{tabular}
        &
        \begin{tabular}{l}
        39.57  {\scriptsize \textcolor{gray}{[37.9, 41.1]}}\\
        \textbf{44.68 {\scriptsize \textcolor{gray}{[43.29, 46.55]}}}\\
        35.51 {\scriptsize \textcolor{gray}{[34.09, 37.89]}}\\
        \textbf{28.38 {\scriptsize \textcolor{gray}{[26.99, 29.94]}}}
        \end{tabular}
        \\
        \hline
        
        Cardiac
        &
        \begin{tabular}{l}
        F1-Score \\
        Precision\\
        Recall \\
        RougeL
        \end{tabular}
        &
        \begin{tabular}{l}
        34.01 {\scriptsize \textcolor{gray}{[31.09, 36.71]}}\\
        21.92 {\scriptsize \textcolor{gray}{[20.87, 23.39]}}\\
        \textbf{75.94 {\scriptsize \textcolor{gray}{[71.1, 81.18]}}} \\
        11.48 {\scriptsize \textcolor{gray}{[9.69, 12.88]}}
        \end{tabular}
        &
        \begin{tabular}{l}
        \textbf{35.69  {\scriptsize \textcolor{gray}{[31.37, 41.0]}}}\\
        35.8 {\scriptsize \textcolor{gray}{[30.59, 39.75]}}\\
        35.59 {\scriptsize \textcolor{gray}{[33.01, 41.29]}}\\
        19.57 {\scriptsize \textcolor{gray}{[17.97, 23.44]}}
        \end{tabular}
        &
        \begin{tabular}{l}
        33.78  {\scriptsize \textcolor{gray}{[30.29, 37.11]}}\\
        28.5 {\scriptsize \textcolor{gray}{[25.48, 33.13]}}\\
        41.45 {\scriptsize \textcolor{gray}{[36.92, 43.22]}}\\
        20.23 {\scriptsize \textcolor{gray}{[17.69, 22.23]}}
        \end{tabular}
        &
        \begin{tabular}{l}
        35.34 {\scriptsize \textcolor{gray}{ [33.47, 39.86]}}\\
        \textbf{40.91 {\scriptsize \textcolor{gray}{[35.7, 44.57]}}}\\
        31.1 {\scriptsize \textcolor{gray}{[27.64, 34.39]}}\\
        \textbf{22.41 {\scriptsize \textcolor{gray}{[20.28, 24.11]}}}
        \end{tabular}
        \\
        \hline
        
        Gastrointestinal
        &
        \begin{tabular}{l}
        F1-Score \\
        Precision\\
        Recall \\
        RougeL
        \end{tabular}
        &
        \begin{tabular}{l}
        33.51 {\scriptsize \textcolor{gray}{[32.68, 34.56]}}\\
        21.61 {\scriptsize \textcolor{gray}{[21.31, 23.06]}}\\
        \textbf{74.58 {\scriptsize \textcolor{gray}{[73.23, 75.98]}}} \\
        11.98 {\scriptsize \textcolor{gray}{[11.61, 12.74]}}
        \end{tabular}
        &
        \begin{tabular}{l}
        37.78 {\scriptsize \textcolor{gray}{ [35.85, 38.88]}}\\
        40.13 {\scriptsize \textcolor{gray}{[38.07, 42.2]}}\\
        35.69 {\scriptsize \textcolor{gray}{[34.2, 38.42]}}\\
        24.23 {\scriptsize \textcolor{gray}{[22.85, 25.44]}}
        \end{tabular}
        &
        \begin{tabular}{l}
        32.93  {\scriptsize \textcolor{gray}{[31.99, 33.66]}}\\
        27.89 {\scriptsize \textcolor{gray}{[26.4, 28.75]}}\\
        40.19 {\scriptsize \textcolor{gray}{[38.34, 41.73]}}\\
        22.45 {\scriptsize \textcolor{gray}{[21.56, 23.01]}}
        \end{tabular}
        &
        \begin{tabular}{l}
        \textbf{38.95  {\scriptsize \textcolor{gray}{[37.77, 40.14]}}} \\
        \textbf{42.85 {\scriptsize \textcolor{gray}{[41.3, 43.95]}}} \\
        35.71 {\scriptsize \textcolor{gray}{[34.24, 37.8]}}\\
        \textbf{26.91 {\scriptsize \textcolor{gray}{[25.8, 28.44]}}}
        \end{tabular}
        \\
        \hline
        
        Musculoskeletal
        &
        \begin{tabular}{l}
        F1-Score \\
        Precision\\
        Recall \\
        RougeL
        \end{tabular}
        &
        \begin{tabular}{l}
        33.55 {\scriptsize \textcolor{gray}{[32.86, 35.07]}}\\
        21.47 {\scriptsize \textcolor{gray}{[20.92, 22.39]}}\\
        \textbf{76.77 {\scriptsize \textcolor{gray}{[74.78, 77.8]}}} \\
        12.49 {\scriptsize \textcolor{gray}{[12.05, 12.95]}}
        \end{tabular}
        &
        \begin{tabular}{l}
        40.11 {\scriptsize \textcolor{gray}{ [39.05, 41.79]}}\\
        43.19 {\scriptsize \textcolor{gray}{[42.14, 44.48]}}\\
        37.43 {\scriptsize \textcolor{gray}{[35.74, 38.19]}}\\
        29.67 {\scriptsize \textcolor{gray}{[29.13, 30.21]}}
        \end{tabular}
        &
        \begin{tabular}{l}
        36.5  {\scriptsize \textcolor{gray}{[34.89, 37.83]}}\\
        31.21 {\scriptsize \textcolor{gray}{[30.12, 32.56]}}\\
        43.96 {\scriptsize \textcolor{gray}{[42.4, 45.4]}}\\
        26.54 {\scriptsize \textcolor{gray}{[25.81, 27.16]}}
        \end{tabular}
        &
        \begin{tabular}{l}
        \textbf{40.84  {\scriptsize \textcolor{gray}{[39.51, 42.38]}}} \\
        \textbf{44.43 {\scriptsize \textcolor{gray}{[43.59, 46.67]}}} \\
        37.78 {\scriptsize \textcolor{gray}{[36.29, 39.3]}}\\
        \textbf{30.18 {\scriptsize \textcolor{gray}{[29.34, 31.86]}}}
        \end{tabular}
        \\
        \hline
        
        Central Nervous System
        &
        \begin{tabular}{l}
        F1-Score \\
        Precision\\
        Recall \\
        RougeL
        \end{tabular}
        &
        \begin{tabular}{l}
        34.18 {\scriptsize \textcolor{gray}{[32.87, 34.51]}}\\
        22.08 {\scriptsize \textcolor{gray}{[21.01, 22.83]}}\\
        \textbf{75.58 {\scriptsize \textcolor{gray}{[73.95, 76.39]}}} \\
        11.58 {\scriptsize \textcolor{gray}{[10.97, 12.17]}}
        \end{tabular}
        &
        \begin{tabular}{l}
        36.62  {\scriptsize \textcolor{gray}{[35.55, 37.8]}}\\
        41.13 {\scriptsize \textcolor{gray}{[39.76, 42.61]}}\\
        33.01 {\scriptsize \textcolor{gray}{[30.63, 34.1]}}\\
        25.41 {\scriptsize \textcolor{gray}{[23.82, 26.16]}}
        \end{tabular}
        &
        \begin{tabular}{l}
        33.46  {\scriptsize \textcolor{gray}{[31.95, 34.96]}}\\
        29.46 {\scriptsize \textcolor{gray}{[28.19, 31.33]}}\\
        38.72 {\scriptsize \textcolor{gray}{[36.46, 40.25]}}\\
        22.37 {\scriptsize \textcolor{gray}{[21.04, 23.27]}}
        \end{tabular}
        &
        \begin{tabular}{l}
        \textbf{37.65 {\scriptsize \textcolor{gray}{[36.1, 40.25] }}} \\
        \textbf{41.07 {\scriptsize \textcolor{gray}{[39.27, 43.37]}}} \\
        34.74 {\scriptsize \textcolor{gray}{[32.79, 37.83]}}\\
        \textbf{24.95 {\scriptsize \textcolor{gray}{[23.94, 26.68]}}}
        \end{tabular}\\
        \hline
        
        Breast
        &
        \begin{tabular}{l}
        F1-Score \\
        Precision\\
        Recall \\
        RougeL
        \end{tabular}
        &
        \begin{tabular}{l}
        31.02 {\scriptsize \textcolor{gray}{[28.96, 32.34]}}\\
        19.26 {\scriptsize \textcolor{gray}{[17.61, 20.71]}}\\
        \textbf{79.64 {\scriptsize \textcolor{gray}{[75.53, 82.85]}}} \\
        9.11 {\scriptsize \textcolor{gray}{[8.41, 9.84]}}
        \end{tabular}
        &
        \begin{tabular}{l}
        44.99  {\scriptsize \textcolor{gray}{[41.83, 49.87]}}\\
        43.87 {\scriptsize \textcolor{gray}{[38.35, 62.14]}}\\
        46.18 {\scriptsize \textcolor{gray}{[35.35, 54.97]}}\\
        27.99 {\scriptsize \textcolor{gray}{[22.04, 38.85]}}
        \end{tabular}
        &
        \begin{tabular}{l}
        36.42  {\scriptsize \textcolor{gray}{[31.85, 42.53]}}\\
        29.93 {\scriptsize \textcolor{gray}{[26.63, 32.68]}}\\
        46.52 {\scriptsize \textcolor{gray}{[39.93, 51.5]}}\\
        25.09 {\scriptsize \textcolor{gray}{[22.12, 28.86]}}
        \end{tabular}
        &
        \begin{tabular}{l}
        \textbf{45.39  {\scriptsize \textcolor{gray}{[34.85, 49.99]}}} \\
        \textbf{45.08 {\scriptsize \textcolor{gray}{[38.27, 50.59]}}} \\
        45.7 {\scriptsize \textcolor{gray}{[41.12, 53.0]}}\\
        \textbf{32.92 {\scriptsize \textcolor{gray}{[29.25, 39.1]}}}
        \end{tabular}\\
        \hline

        Urogenital
        &
        \begin{tabular}{l}
        F1-Score \\
        Precision\\
        Recall \\
        RougeL
        \end{tabular}
        &
        \begin{tabular}{l}
        33.97 {\scriptsize \textcolor{gray}{ [31.68, 36.49]}}\\
        21.5 {\scriptsize \textcolor{gray}{[19.96, 22.91]}}\\
        \textbf{80.94 {\scriptsize \textcolor{gray}{[77.25, 82.03]}}} \\
        10.13 {\scriptsize \textcolor{gray}{[9.32, 11.0]}}
        \end{tabular}
        &
        \begin{tabular}{l}
        38.38  {\scriptsize \textcolor{gray}{[36.1, 41.82]}}\\
        40.76 {\scriptsize \textcolor{gray}{[37.62, 46.24]}}\\
        36.27 {\scriptsize \textcolor{gray}{[29.97, 39.91]}}\\
        26.95 {\scriptsize \textcolor{gray}{[24.36, 30.56]}}
        \end{tabular}
        &
        \begin{tabular}{l}
        34.14  {\scriptsize \textcolor{gray}{[30.74, 37.34]}}\\
        28.26 {\scriptsize \textcolor{gray}{[24.71, 29.36]}}\\
        43.11 {\scriptsize \textcolor{gray}{[34.54, 46.95]}}\\
        21.86 {\scriptsize \textcolor{gray}{[18.85, 22.79]}}
        \end{tabular}
        &
        \begin{tabular}{l}
        \textbf{40.55 {\scriptsize \textcolor{gray}{[36.93, 44.36] }}} \\
        \textbf{41.81 {\scriptsize \textcolor{gray}{[36.55, 44.33]}}} \\
        39.36 {\scriptsize \textcolor{gray}{[35.77, 45.5]}}\\
        \textbf{27.78 {\scriptsize \textcolor{gray}{[23.7, 30.27]}}}
        \end{tabular}\\
        \hline

        Vascular
        &
        \begin{tabular}{l}
        F1-Score \\
        Precision\\
        Recall \\
        RougeL
        \end{tabular}
        &
        \begin{tabular}{l}
        28.9  {\scriptsize \textcolor{gray}{[28.1, 31.06]}}\\
        17.85 {\scriptsize \textcolor{gray}{[14.67, 19.32]}}\\
        \textbf{75.88 {\scriptsize \textcolor{gray}{[72.83, 77.13]}}} \\
        9.84 {\scriptsize \textcolor{gray}{[8.26, 11.48]}}
        \end{tabular}
        &
        \begin{tabular}{l}
        34.66 {\scriptsize \textcolor{gray}{ [30.18, 40.45]}}\\
        34.55 {\scriptsize \textcolor{gray}{[34.37, 38.27]}}\\
        34.77 {\scriptsize \textcolor{gray}{[33.26, 38.84]}}\\
        23.42 {\scriptsize \textcolor{gray}{[21.51, 26.46]}}
        \end{tabular}
        &
        \begin{tabular}{l}
        33.53  {\scriptsize \textcolor{gray}{[29.83, 38.82]}}\\
        27.27 {\scriptsize \textcolor{gray}{[23.98, 30.64]}}\\
        43.52 {\scriptsize \textcolor{gray}{[41.04, 50.72]}}\\
        21.97 {\scriptsize \textcolor{gray}{[19.85, 25.3]}}
        \end{tabular}
        &
        \begin{tabular}{l}
        \textbf{39.15 {\scriptsize \textcolor{gray}{ [37.03, 44.21]}}} \\
        \textbf{38.58 {\scriptsize \textcolor{gray}{[29.55, 47.74]}}} \\
        39.74 {\scriptsize \textcolor{gray}{[34.97, 47.79]}}\\
        \textbf{24.75 {\scriptsize \textcolor{gray}{[19.9, 33.26]}}}
        \end{tabular}\\
        \hline

        Forensic
        &
        \begin{tabular}{l}
        F1-Score \\
        Precision\\
        Recall \\
        RougeL
        \end{tabular}
        &
        \begin{tabular}{l}
        27.41  {\scriptsize \textcolor{gray}{[24.09, 32.25]}}\\
        16.75 {\scriptsize \textcolor{gray}{[14.17, 19.99]}}\\
        \textbf{75.41 {\scriptsize \textcolor{gray}{[69.56, 85.01]}}} \\
        11.54 {\scriptsize \textcolor{gray}{[7.15, 15.12]}}
        \end{tabular}
        &
        \begin{tabular}{l}
        44.99  {\scriptsize \textcolor{gray}{[38.99, 50.29]}}\\
        46.27 {\scriptsize \textcolor{gray}{[38.47, 58.28]}}\\
        43.78 {\scriptsize \textcolor{gray}{[38.14, 47.7]}}\\
        43.52 {\scriptsize \textcolor{gray}{[34.16, 50.44]}}
        \end{tabular}
        &
        \begin{tabular}{l}
        32.32 {\scriptsize \textcolor{gray}{ [25.5, 36.49]}}\\
        24.38 {\scriptsize \textcolor{gray}{[20.13, 32.99]}}\\
        47.95 {\scriptsize \textcolor{gray}{[37.47, 59.6]}}\\
        28.62 {\scriptsize \textcolor{gray}{[20.44, 42.75]}}
        \end{tabular}
        &
        \begin{tabular}{l}
        \textbf{49.25 {\scriptsize \textcolor{gray}{ [41.88, 57.61]}}} \\
        \textbf{48.72 {\scriptsize \textcolor{gray}{[37.22, 54.07]}}} \\
        49.79 {\scriptsize \textcolor{gray}{[42.04, 56.95]}}\\
        \textbf{38.43 {\scriptsize \textcolor{gray}{[32.25, 43.37]}}}
        \end{tabular}\\
        \hline

        Hepatobiliary
        &
        \begin{tabular}{l}
        F1-Score \\
        Precision\\
        Recall \\
        RougeL
        \end{tabular}
        &
        \begin{tabular}{l}
        \textbf{37.51 {\scriptsize \textcolor{gray}{ [36.44, 40.21]}}}\\
        24.83 {\scriptsize \textcolor{gray}{[20.98, 26.59]}}\\
        \textbf{76.62 {\scriptsize \textcolor{gray}{[72.71, 80.57]}}} \\
        13.45 {\scriptsize \textcolor{gray}{[10.62, 14.65]}}
        \end{tabular}
        &
        \begin{tabular}{l}
        36.9  {\scriptsize \textcolor{gray}{[30.59, 44.76]}}\\
        \textbf{41.25 {\scriptsize \textcolor{gray}{[33.54, 48.13]}}}\\
        33.38 {\scriptsize \textcolor{gray}{[23.55, 46.46]}}\\
        22.51 {\scriptsize \textcolor{gray}{[19.43, 26.85]}}
        \end{tabular}
        &
        \begin{tabular}{l}
        29.94  {\scriptsize \textcolor{gray}{[28.47, 34.41]}}\\
        27.48 {\scriptsize \textcolor{gray}{[26.11, 31.82]}}\\
        32.89 {\scriptsize \textcolor{gray}{[26.1, 46.39]}}\\
        23.02 {\scriptsize \textcolor{gray}{[18.83, 27.28]}}
        \end{tabular}
        &
        \begin{tabular}{l}
        36.57 {\scriptsize \textcolor{gray}{ [30.16, 41.52]}} \\
        39.01 {\scriptsize \textcolor{gray}{[32.36, 39.77]}} \\
        34.43 {\scriptsize \textcolor{gray}{[30.27, 39.71]}}\\
        \textbf{27.15 {\scriptsize \textcolor{gray}{[20.53, 30.1]}}}
        \end{tabular}\\
        \hline       
    \end{tabular}}
    \label{tab:radiopaedia_metrics}
\end{table*}


\section{Examples}\label{apd:third}

\begin{table*}[!ht]
    \centering
    \caption{Abnormality Label Prediction Example}
    \begin{tabular}{|p{50pt}|p{400pt}|}
    \hline
         Prompt& Given the below radiology report:
         
         PREAMBLE: PA AND LATERAL VIEWS OF THE CHEST
         
         INDICATION: Shortness of breath , wheezing on exam. The right hemidiaphragm is elevated. Aside from linear atelectasis in the right base, the lungs are clear. There is no pneumothorax or pleural effusion. Cardiomediastinal contours are normal. The sternal wires are aligned.
         
         IMPRESSION: No evidence of acute cardiopulmonary abnormalities.
         
         What abnormality labels can be applied to these findings?\\
         \hline
        Ground Truth & atelectasis, linear/patchy atelectasis, lung opacity, elevated hemidiaphragm\\
        \hline
        RadPhi-3 & atelectasis, linear/patchy atelectasis, lung opacity, elevated hemidiaphragm\\
        \hline
    \end{tabular}
    
    \label{tab:example_abnormality_label}
\end{table*}

\begin{table*}[!ht]
    \centering
    \caption{QA Comprehension Example}
    \begin{tabular}{|p{50pt}|p{400pt}|}
    \hline
         Prompt& Use the information from the radiology report below to answer the question that follows:
         
         PREAMBLE: PORTABLE CHEST, \_
         
         COMPARISON: \_ study.
         
         FINDINGS: The patient is status post median sternotomy and coronary bypass surgery. Heart remains enlarged, and is accompanied by pulmonary vascular congestion. Interval improved aeration at both lung bases with improving atelectasis and decreasing pleural effusions. No new areas of consolidation within either lung.
         
         what level is the pleural effusion?  \\
         \hline
        Ground Truth & decreasing\\
        \hline
        Phi3-mini-4k-instruct& The pleural effusion is described as \"decreasing,\" which implies that it is present but is lessening in severity. However, the report does not specify the exact level of the pleural effusion. To determine the level, a radiologist would typically look for the lowest point of the fluid collection on a chest X-ray or CT scan. Since the report does not provide a specific location, it is not possible to give a precise level without additional imaging details.\\
        \hline
        RadPhi-2 & decreasing\\
        \hline
        RadPhi-3 & decreasing\\
        \hline
    \end{tabular}
    
    \label{tab:example_qacomprehension}
\end{table*}

\begin{table*}[!ht]
    \centering
    \footnotesize
    \caption{Radiology Question Answering Example}
    \begin{tabular}{|p{50pt}|p{400pt}|}
    \hline
         Prompt& What is the epidemiology of abdominal aortic aneurysm?\\
         \hline
        Ground Truth & The prevalence of rupture increases with age, males are more commonly affected than females, prevalence is almost 10\% in people over 65 years old, and it is the 10th commonest cause of death in Western populations.\\
        \hline
        Phi-3-mini-4k-instruct&Abdominal aortic aneurysm (AAA) is a localized enlargement of the abdominal aorta such that the diameter is greater than 3 cm or more than 50\% larger than normal diameter. The epidemiology of AAA varies by region, but it is more common in men than in women, with a male-to-female ratio of approximately 4:1. The prevalence increases with age, with the majority of cases occurring in individuals over the age of 65. Smoking is a significant risk factor, and the prevalence of AAA is higher in smokers compared to non-smokers. Other risk factors include hypertension, atherosclerosis, family history of AAA, and male gender. The incidence of AAA is estimated to be about 2 to 4 per 1000 person-years in the general population. In some countries, such as the United Kingdom, screening programs for AAA have been implemented, which have led to increased detection rate.\\
        \hline
        GPT-4& The epidemiology of abdominal aortic aneurysm (AAA) shows a prevalence of around 2-8\% among older adults, especially in men over 65. Risks include age, smoking, hypertension, and a family history of AAA. Incidence is declining due to decreased smoking rates and improved hypertension control.\\
        \hline
        RadPhi-3 &  Abdominal aortic aneurysms are more common in men than in women. The prevalence increases with age, and the majority of aneurysms are diagnosed in patients over 65 years of age.\\
        \hline
    \end{tabular}
    \label{tab:example_radqa}
\end{table*}

\begin{table*}[!ht]
    \centering
     \caption{Padchest Label Prediction Example}
    \begin{tabular}{|p{50pt}|p{400pt}|}
    \hline
         Prompt& 

         Considerando el siguiente informe radiol´ogico,

cambi pulmonar cronic con tract fibros lsd . aument densid parec extrapleural call osea fals imag nivel medi hemitorax derech . cit complet estudi con tc torax .

¿ cu´ales son las posibles anomal´ıas o etiquetas de dispositivos para
los hallazgos en el informe radiol´ogico ?\\
         \hline
        Ground Truth & increased density,fibrotic band,chronic changes\\
        \hline
        RadPhi-3 &chronic changes, fibrotic band,increased density\\
        \hline
    \end{tabular}
   
    \label{tab:example_padchest}
\end{table*}

\begin{table*}[!ht]
    \centering
    \caption{Impression Prediction Example}
    \begin{tabular}{|p{50pt}|p{400pt}|}
    \hline
         Prompt& Given the findings from a radiology report:
         
         An extensive right hilar lung mass is associated with radiation fibrosis, better delineated on CT \_. An additional component of postobstructive pneumonia may be present. Retrocardiac opacity, left pleural effusion, and left plueral thickening are also new. No pneumothorax is present.
         
         The findings from the radiology report above need an impression. Write one.  \\
         \hline
        Ground Truth & Large right hilar lung mass and radiation fibrosis are present. There is an opacity in the left retrocardiac region. There is a small left effusion and thickening of the pleura.\\
        \hline
        Phi-3-mini-4k-instruct & The patient presents with a significant right hilar mass suggestive of a neoplastic process, likely secondary to prior radiation therapy, as indicated by the presence of radiation fibrosis. The mass is more clearly defined on CT imaging. There is also a suspicion of postobstructive pneumonia, which may be contributing to the patient's respiratory symptoms. New findings include a retrocardiac opacity, which could represent a mass or consolidation, a left pleural effusion, and left pleural thickening. These findings warrant further investigation, including a possible biopsy of the hilar mass and evaluation of the pleural effusion. No evidence of pneumothorax is noted.\\
        \hline
        RadPhi2 & Right hilar mass is present. There is a component of postobstructive pneumonia. Left lower lobe opacity is present, likely due to a combination of pleural effusion and atelectasis.\\
        \hline
        RadPhi-3 & Right hilar mass is present. There is radiation fibrosis. There is a left lower lobe pneumonia. There is a left pleural effusion.\\
        \hline
    \end{tabular}
    
    \label{tab:example_impression_prediction}
\end{table*}

\begin{table*}[!ht]
    \centering
    \caption{Cleanup Radiology Text Example}
    \begin{tabular}{|p{50pt}|p{400pt}|}
    \hline
         Prompt& Given the findings from a radiology report:
         
         The cardiomediastinal and hilar silhouettes remain stable. There are bilateral upper lung reticular and nodular opacities with associated volume loss, stable compared to the prior study. The lungs are otherwise clear with no focal consolidation. There is no pleural effusion, pulmonary edema, or pneumothorax. The osseous structures are unremarkable.
         
         Update the findings such that each sentence in the output describes a finding or observation about the image. Remove any mention of change of an observation and just state its presence. Do not include any follow-up suggestions or advice, and avoid mentioning any doctor names or other noisy details. \\
         \hline
        Ground Truth & There are bilateral upper lung reticular and nodular opacities with associated volume loss. The lungs are otherwise clear with no focal consolidation. There is no pleural effusion, pulmonary edema, or pneumothorax. The osseous structures are unremarkable.\\
        \hline
        RadPhi-2& Bilateral upper lung reticular and nodular opacities with associated volume loss are present. The lungs are clear with no focal consolidation. There is no pleural effusion, pulmonary edema, or pneumothorax. The osseous structures are unremarkable.\\
        \hline
        RadPhi-3 & Bilateral upper lung reticular and nodular opacities with associated volume loss are present. The lungs are clear with no focal consolidation. There is no pleural effusion, pulmonary edema, or pneumothorax. The osseous structures are unremarkable.\\
        \hline
    \end{tabular}
    
    \label{tab:example_cleanup_radiology}
\end{table*}

\begin{table*}[!ht]
    \centering
    \caption{Impression Extraction Example}
    \begin{tabular}{|p{50pt}|p{400pt}|}
    \hline
         Prompt& Given the radiology report below:
         
         INDICATION: History of sarcoidosis, now status post laparoscopic Nissen on post-op day \# 2 with respiratory desaturations, here to evaluate for recurrent pneumonia.
         
         COMPARISON: Chest radiograph, last performed on \_.
         
         TECHNIQUE: PA and lateral radiographs of the chest.
         
         FINDINGS: The inspiratory lung volumes are decreased from the most recent prior study. Increased opacification of the bilateral bases is likely reflective of atelectasis in the post-operative setting of low lung volumes. Prominent lung markings raise the possibility of chronic lung disease. Small bilateral pleural effusions are present. No pneumothorax is detected. The cardiac silhouette is likely within normal limits allowing for low lung volumes. No overt pulmonary edema is present. Calcified hilar and mediastinal lymph nodes are re-demonstrated, compatible with sequela of known sarcoidosis. Multiple healed right posterior rib fractures are again noted.
         
         IMPRESSION: 1. Bibasilar opacities most likely represent atelectasis in the post-operative setting of low lung volumes. Small pleural effusions. 2. Stable calcified hilar and mediastinal lymph nodes compatible with known sarcoidosis. 3. Suggestion of chronic lung disease.
         
         Output the impression of the radiology report. Do not mention any doctor names, noisy details, or follow-up advice or changes in observations. \\
         \hline
        Ground Truth & Bibasilar opacities most likely represent atelectasis. Small pleural effusions are present. Calcified hilar and mediastinal lymph nodes compatible with sarcoidosis. There is suggestion of chronic lung disease.\\
        \hline
        RadPhi-3 & Bibasilar opacities are present, likely representing atelectasis in the post-operative setting of low lung volumes. Small pleural effusions are observed. Calcified hilar and mediastinal lymph nodes are present, compatible with known sarcoidosis. There is suggestion of chronic lung disease.\\
        \hline
    \end{tabular}
    
    \label{tab:example_extraction}
\end{table*}

\begin{table*}[!ht]
    \centering
    \caption{Tubes, Lines and Device Label Prediction Example}
    \begin{tabular}{|p{50pt}|p{400pt}|}
    \hline
         Prompt& Given the below radiology report:
         
         EXAMINATION: CHEST (PORTABLE AP)
         
         INDICATION: \_ year old man with very reduced EF, s/p Respiratory and PEA arrest and intubation \_ now with OG tube. // OG tube placement OG tube placement
         
         COMPARISON: Prior chest radiographs \_ through \_.
         
         IMPRESSION: More dependent distribution of right pleural effusion is due to more positioning of the patient. Previously questioned new right upper lobe consolidation is smaller than assumed. Moderate cardiomegaly and pulmonary vascular congestion persist. There is probably minimal pulmonary edema. ET tube in standard placement. Right jugular line ends in the upper SVC. Nasogastric tube ends at the diaphragm and would need to be advanced at least 15 cm to move all side ports into the stomach.
         
         What are the tubes and lines or devices that are mentioned in the radiology report?\\
         \hline
        Ground Truth & enteric tube, endotracheal tube, ij line\\
        \hline
        RadPhi-3 & ij line, endotracheal tube, enteric tube\\
        \hline
    \end{tabular}
    
    \label{tab:example_tubesnlines_label}
\end{table*}

\begin{table*}[!ht]
    \centering
    \caption{NLI Example}
    \begin{tabular}{|p{50pt}|p{400pt}|}
    \hline
         Prompt& In this task, you are given a premise and hypothesis.
         
         The task is to classify them into three categories: 'positive' if the hypothesis supports the premise, 'negated' if it opposes the premise, and 'neutral' if it neither supports nor opposes it.
         
         Premise: A recent TEE showed severe aortic stenosos with an aortic valve area of 0.7cm2.
         
         Hypothesis:  Patient has a normal echo\\
         \hline
        Ground Truth & negated\\
        \hline
        RadPhi-3 & negated\\
        \hline
    \end{tabular}
    
    \label{tab:example_RADNLI}
\end{table*}

\begin{table*}[!ht]
    \centering
    \caption{RadGraph2 - Temporal Summary Example}
    \begin{tabular}{|p{50pt}|p{400pt}|}
    \hline
         Prompt& Given a radiology report generate a change summary
         
         Report:
         
         FINAL REPORT CHEST RADIOGRAPH PERFORMED ON \_\_\_ COMPARISON : Prior exam from \_\_\_ . CLINICAL HISTORY : Cirrhosis with confusion , question pneumonia . FINDINGS : AP portable upright chest radiograph was provided . As seen previously , there are bilateral pleural effusions slightly increased on the left and stable on the right . The subjacent opacities in the lower lungs could represent atelectasis versus pneumonia . The upper lungs are well aerated . The cardiomediastinal silhouette is stable . Bony structures are intact . TIPS partially imaged in the right upper quadrant . IMPRESSION : Bilateral pleural effusions , subjacent consolidations , may represent atelectasis versus pneumonia .\\
         \hline
        Ground Truth & Change summary of medical devices:
        
        Change summary of medical conditions:
        
        \quad Worsened:
        
        \quad  \quad As seen previously , there are bilateral pleural effusions slightly increased on the left and stable on the right .
        
        No change summary:
        
        \quad Stable:
        
        \quad \quad As seen previously , there are bilateral pleural effusions slightly increased on the left and stable on the right .
        
        \quad  \quad  The cardiomediastinal silhouette is stable .\\
        \hline
        RadPhi-3 & Change summary of medical devices:
        
        Change summary of medical conditions:
        
        \quad Worsened:
        
        \quad \quad As seen previously, there are bilateral pleural effusions slightly increased on the left and stable on the right.
        
        No change summary:
        
        \quad Stable:
        
        \quad \quad The cardiomediastinal silhouette is stable.
        
        \quad \quad As seen previously, there are bilateral pleural effusions slightly increased on the left and stable on the right.\\
        \hline
    \end{tabular}
    
    \label{tab:example_temporal_extraction}
\end{table*}

\begin{table*}[!ht]
    \centering
    \caption{CheXpert Section Extraction Example}
    \begin{tabular}{|p{50pt}|p{400pt}|}
    \hline
         Prompt& Given the radiology report below:
         NARRATIVE:
         RADIOGRAPHIC EXAMINATION OF THE CHEST: 2002 january 26th
         CLINICAL HISTORY: 53 years of age, Female, postop.
         COMPARISON: Same day radiograph
         PROCEDURE COMMENTS: Single view of the chest. 
         FINDINGS:
         The transesophageal echo probe has been removed. A new enteric tube is present. There is otherwise unchanged positioning of supportive medical devices.
         Mild pulmonary edema and cardiomegaly.
         Left basilar opacity. No pneumothorax.
         No acute bony abnormalities are noted.
         IMPRESSION:
         1.  Residual mild pulmonary edema and left basilar opacity.
         "Physician to Physician Radiology Consult Line: (898) 940-4661"
         Signed
         ACCESSION NUMBER:
         04240392 This report has been anonymized. All dates are offset from the actual dates by a fixed interval associated with the patient.
         
         Extract all the important sections from it for better interpretability.\\
         \hline
        Ground Truth & Clinical History:
        
        \quad 53 years of age, Female, postop.
        
        Comparison:
        
        \quad Same day radiograph
        
        Technique:
        
        \quad N/A
        
        Procedure Comments:
        
        \quad Single view of the chest. 
        
        Findings:
        
        \quad The transesophageal echo probe has been removed.  A new enteric tube is present. There is otherwise unchanged positioning of supportive medical devices.  Mild pulmonary edema and cardiomegaly. Left basilar opacity.No pneumothorax. No acute bony abnormalities are noted.
        
        Impression:
        
        \quad 1.  Residual mild pulmonary edema and left basilar opacity."Physician to Physician Radiology Consult Line: (898) 940-4661"
        
        Signed
        
        End Of Impression:
        
        \quad N/A
        
        Summary:
        
        \quad N/A\\
        \hline
        RadPhi-3 & Clinical History:
        
        \quad 53 years of age, Female, postop.
        
        Comparison:
        
        \quad Same day radiograph
        
        Technique:
        
        \quad N/A
        
        Procedure Comments:
        
        \quad Single view of the chest. 
        
        Findings:
        
        \quad The transesophageal echo probe has been removed.  A new enteric tube is present. There is otherwise unchanged positioning of supportive medical devices.  Mild pulmonary edema and cardiomegaly. Left basilar opacity.No pneumothorax. No acute bony abnormalities are noted.
        
        Impression:
        
        \quad 1.  Residual mild pulmonary edema and left basilar opacity."Physician to Physician Radiology Consult Line: (898) 940-4661"
        
        Signed
        
        End Of Impression:
        
        \quad N/A
        
        Summary:
        
        \quad N/A\\
        \hline
    \end{tabular}
    \label{tab:example_chexpert_section_extraction}
\end{table*}

\begin{table*}[!ht]
    \centering
    \caption{Temporal Summary Example}
    \begin{tabular}{|p{50pt}|p{400pt}|}
    \hline
         Prompt& Given a current radiology report and a previous report of the same patient, generate a change summary
         Current Report:
         FINAL REPORT
         SINGLE AP PORTABLE VIEW OF THE CHEST
         REASON FOR EXAM:  AVR, assess for occlusions.
         COMPARISON:  Comparison is made with prior study of \_\_\_.
         Cardiomegaly is stable.  Vascular congestion has improved, now is mild.  Right lower lobe atelectasis has increased.  Aeration of the left hemidiaphragm has minimally improved.  Left IJ catheter tip is in the mid SVC.  There is no evidence of pneumothorax.  Small bilateral pleural effusions are unchanged.
         Previous Report:
         FINAL REPORT
         CHEST RADIOGRAPH
         TECHNIQUE:  Single portable upright radiograph of the chest was compared with prior radiograph from \_\_\_.
         FINDINGS:
         Since prior radiograph from \_\_\_, the mediastinal drain tube has been removed.  There is no pneumothorax. Both lung volumes are very low. Bilateral, right side more than left side,  moderate pulmonary edema has improved.  Widened cardiomediastinal silhouette is more than it was on \_\_\_; however, this appearance could be exacerbation from low lung volumes.  Patient is status post median sternotomy with intact sternal sutures.\\
         \hline
        Ground Truth & Diseases\_Change\_Summary:
        
        \quad New:
        
        \quad Resolved:
        
        \quad Stable:
        
        \quad \quad Cardiomegaly
        
        \quad \quad Small bilateral pleural effusions
        
        \quad Improved:
        
        \quad \quad Vascular congestion
        
        \quad \quad Aeration of the left hemidiaphragm
        
        \quad Worsened:
        
        \quad \quad Right lower lobe atelectasis
        
        \quad Negatives:
        
        \quad \quad There is no evidence of pneumothorax
        
        Tubes\_Lines\_Change\_Summary:
        
        \quad New:
        
        \quad \quad Left IJ catheter tip is in the mid SVC
        
        \quad Removed:
        
        \quad Unchanged:
        
        \quad Changed:
        
        Recommendations:\\
        \hline
        RadPhi-3 & Diseases\_Change\_Summary:
        
        \quad New:
        
        \quad Resolved:
        
        \quad Stable:
        
        \quad \quad Cardiomegaly
        
        \quad \quad Small bilateral pleural effusions
        
        \quad Improved:
        
        \quad \quad Vascular congestion
        
        \quad \quad Aeration of the left hemidiaphragm
        
        \quad Worsened:
        
        \quad \quad Right lower lobe atelectasis
        
        \quad Negatives:
        
        \quad \quad No evidence of pneumothorax
        
        Tubes\_Lines\_Change\_Summary:
        
        New:
        
        \quad \quad Left IJ catheter tip is in the mid SVC
        
        Removed:
        
        \quad \quad Mediastinal drain tube
        
        \quad Unchanged:
        
        \quad Changed:
        
        \quad Recommendations:\\
        \hline
    \end{tabular}
    
    \label{tab:temporal_extraction_example}
\end{table*}

\clearpage
\section{Prompts}\label{apd:fourth}

\textbf{Prompt Design - Temporal Summary Extraction}

\label{sec:temporal_extraction_summary_prompt}
\noindent\fbox{%
    \parbox{\textwidth}{%
\scriptsize Given a current radiology report and a previous report of the same patient, generate a change summary with the following details:

Change Summary for Diseases:
List of new conditions reported in the current report that did not exist before or list of conditions that were existing before and which were either completely resolved, improved, worsened , or unchanged (stable), all according to the current report. Conditions are considered stable if they are mentioned as stable or unchanged in the current report. Resolved conditions should be explicitly mentioned as fully resolved or cleared in the current report, and it should not include conditions that are just absent in the current report. Improved conditions should be reported under improvements and not as resolved unless fully resolved. New conditions are the conditions newly mentioned in the current report which were not mentioned in the previous report.
 Improvement or worsened conditions should also be explicitly stated as such in the current report. Negative findings are extracted under the negative category. However, mentions like "there is no new consolidation" mean the condition is unchanged or stable. Extract the progression category as new/resolved/stable/worsened/improved/negative, along with the corresponding progression description. This section should not include the changes related to devices, tubes, or lines.

Tubes and Lines Change Summary:
List of findings related to any devices, tubes, or lines in the patient's body. This could be a new device or a line introduced into the patient's body which was not present before, or it was present before but has been removed now, or the device is present in both cases but the position of the device or line may have changed or remained unchanged. To be reported as unchanged, the device or line should be mentioned in both reports and described as stable or unchanged in the current report. To be reported as changed, the device or line/tube should be mentioned in both reports, and its position  should have changed according to the current report (this does not refer to a recommendation for change but to a change in the placement of the device compared to the previous report). Extract the placement category as new/removed/changed/stable, with the corresponding placement description of the device/line/tube in the body. If there are recommendations regarding device/line/tube placement, they should be extracted separately under recommendations. If any device/tube/line is mentioned as "in place" or "in standard placement," it just means that it is in the right position and should not be confused with the definition of stable or unchanged, which means that the device was mentioned in both reports and its position has not changed.

Extract in the following JSON format:

\{
  "Diseases\_Change\_Summary": \{

    "New": \textless List of Findings that are new according to the current report.
    They should not be present in the previous report. \textgreater or [],

    "Resolved": \textless List of Findings that are mentioned as fully
    resolved or cleared in the current report. \textgreater or [],

    "Stable": \textless List of Findings that are unchanged or stable
    according to the current report. \textgreater or [],

    "Improved": \textless List of Findings that have been explicitly
    reported as improved according to the current report. \textgreater or [],

    "Worsened": \textless List of Findings that have been explicitly reported
    as worsened according to the current report. \textgreater or [],

    "Negatives": \textless List of conditions mentioned negatively in the current report. \textgreater or []

  \}, 

 "Tubes\_Lines\_Change\_Summary": \{

    "New": \textless List of placement descriptions of new devices in the body according to the current report. \textgreater or [],

    "Removed": \textless List of Devices/Lines that have been removed according to the current report. \textgreater or [],

    "Unchanged": \textless List of placement descriptions of devices in the body whose position is unchanged according to the current report. \textgreater or [],

    "Changed": \textless List of placement descriptions of devices in the body     whose position has changed according to the current report. \textgreater
    or [],

    "Recommendations": \textless List of placement recommendations
    on the positioning of devices in the body. \textgreater or []

  \}

\}

If there are no extractions for a particular category, just return an empty list for that category. No findings should overlap across categories. Additionally, do not miss reporting any finding in the current report. It should be covered under one of the two section categories (do not include symptoms of the patient, but only radiographic findings with respect to the current report). The changes are written only with respect to findings and tube/line mentions in the current report. The previous report may also have changes with respect to its prior report,  which should be ignored.

}%
}

\textbf{Prompt Design - Cleanup Radiology Text}
\label{sec:prompt_cleanup_radiology_text}

\noindent\fbox{%
    \parbox{\textwidth}{%
\scriptsize
Given the below findings from a radiology report. 
Paraphrase the findings into individual sentences without mention of priors.

\{FINDINGS\}     
}%
}

\textbf{Prompt Design - Radiopaedia Question Answering Dataset}

\label{sec:prompt_radiopaedia}
\noindent\fbox{%
    \parbox{\textwidth}{
    \scriptsize
    Given a radiology article in an html format extract the content of the entire article as longform question and answer pairs using the format mentioned below. The article content is structured under various headings. Create question answer pair for every heading of the article framing the question using the heading and answer as the content under the heading. If there are sub-headings or bullet points, then create question answer pair for them as well if it cannot be included in the main heading. If there are descriptions as sub-bullet points include them as well. All content of the article should be included in one of the question answer pairs. The questions and answers should strictly be sourced from the article itself.

Question-Answer format: 
[{"question": question from the article, "answer": answer from the article}]
    
    }

}

\begin{table*}[ht]
\label{sec:prompt_radiology_report_tasks}
\centering
\caption{Prompt Design - Instruction Tuning Dataset for Radiology Reports.}
\begin{tabular}{|p{60pt} |p{80pt}| p{300pt}|}
\hline
Source&Task Name&Prompt\\
\hline

Mimic-CXR &Findings/Impression Extraction&
Given the radiology report below:

FULL REPORT

Output the findings/impression of the radiology report. Each sentence in the output should describe an observation or a finding about the image. Do not mention any changes in observations, follow-up suggestions, doctor names, or noisy details. \\
\hline

Mimic-CXR &Cleanup Report Text&Given the text from a radiology report:

RADIOLOGY REPORT TEXT

Update the impressions or findings such that each sentence in the output describes an impression or observation about the image. Remove any mention of change of an observation and just state its presence. Do not include any follow-up suggestions or advice, and avoid mentioning any doctor names or other noisy details. 

\\
\hline
Mimic-CXR &Impression Prediction&Given the findings from a radiology report:

{FINDINGS}

Based on the above findings from a radiology report, write an impression.\\
\hline
Chest-Imagenome &Abnormality Labels&Given the below radiology report: 

{FULL REPORT}

What abnormality labels can be tagged to these findings?\\

\hline
Medical-Diff-VQA&QA Comprehension&Answer the question using the radiology report below as context:

FULL REPORT

Question:\\
\hline
CheXpert Plus&Chexpert Section Extraction& Given the radiology report below:
REPORT TEXT
Extract all the important sections from it for better interpretability.
\\
\hline
Mimic-CXR &Temporal Summary & 
Given a current radiology report and a previous report of the same patient, generate a change summary.

Current Report: CURRENT\_REPORT

Previous Report: PREVIOUS\_REPORT
\\
\hline
MedNLI \& RadNLI &Natural Language Inference& In this task, you are given a premise and hypothesis. 

The task is to classify them into three categories: 'positive' if the hypothesis supports the premise, 'negated' if it opposes the premise, and 'neutral' if it neither supports nor opposes it.

Premise: 

Hypothesis:  
\\
\hline
Chest-Imagenome&Tubes, Lines and Devices Labels &
Given the below radiology report:

FULL\_REPORT

The radiology report above has some tubes and lines or devices. What are they?
\\
\hline
Padchest & Label Prediction & Considerando el siguiente informe radiológico,

FULL\_REPORT

¿ cuáles son las posibles anomalías o etiquetas de dispositivos para los hallazgos en el informe radiológico ?
\\
\hline
\end{tabular}
\label{tab:prompt_design}
\end{table*}

\clearpage

\section{Generalization to New Tasks - Examples}
\label{sec: generalization}
\begin{table*}[!ht]
    \centering
    \caption{Temporal Sentence Similarity Task - MS-CXR-T Example-1}
    \begin{tabular}{|p{50pt}|p{400pt}|}
    \hline
         Prompt& You are a radiologist. Assess whether two sentences are describing the same meaning (paraphrase) or different meaning (different) regarding the change information. Reply with paraphrase or different
         
         - - INPUT
         
         Sentence1: right mid lung opacity has slightly improved.
         
         Sentence2: right mid lung opacity has slightly worsened.
         
         ANSWER:\\
         Ground Truth& different\\
         \hline
         Phi3-mini-4k-instruct&different\\
         \hline
        Rad-Phi2-Instruct& improved\\
        \hline
         Rad-Phi3-Instruct& different\\
        \hline

    \end{tabular}
    
    \label{tab:example1_temporal_summary_mscxr}
\end{table*}

\begin{table*}[!ht]
    \centering
    \caption{Temporal Sentence Similarity Task - MS-CXR-T Example-2}
    \begin{tabular}{|p{50pt}|p{400pt}|}
    \hline
         Prompt& You are a radiologist. Assess whether two sentences are describing the same meaning (paraphrase) or different meaning (different) regarding the change information. Reply with paraphrase or different.
         
         - - INPUT
         
         Sentence1: status post right thoracentesis with interval decrease in right pleural effusion, which is now trace in size.
         
         Sentence2: status post right thoracentesis with interval increase in right pleural effusion, which is now trace in size.
         
         ANSWER:\\
         Ground Truth& different\\
         \hline
         Phi3-mini-4k-instruct&different\\
         \hline
        Rad-Phi2-Instruct& no\\
        \hline
         Rad-Phi3-Instruct& different\\
        \hline

    \end{tabular}
    
    \label{tab:example2_temporal_summary_mscxr}
\end{table*}

\begin{table*}[!ht]
    \centering
    \caption{RAD-QA Dataset example}
    \begin{tabular}{|p{50pt}|p{400pt}|}
    \hline
         Prompt& 
Given below the question: Are thre any changes in the diffuse convalescent opacities in the lungs?

and the radiology report:

FINAL REPORT 

INDICATION:  Neutropenia and Pneumocystis pneumonia.

CHEST X-RAY, PORTABLE AP:

COMPARISON: \_

Endotracheal tube is properly positioned with the tip at the thoracic inlet. A left subclavian central venous line is again demonstrated with tip in the mid- superior vena cava.  No pneumothorax.  There has been interval improvement in the bilateral ground glass, patchy opacities.  Nasogastric tube is also seen properly positioned in the stomach. 

Extract the full sentence from the report that can answer the question. Don't answer the question, just extract the part of the radiology report which can answer the question.If there is no sentence that can be used to answer the question return Not possible to extract.\\
\hline
Ground Truth & improvement in the bilateral ground glass, patchy opacities\\
         \hline
        Phi3-mini-4k-instruct &  There has been interval improvement in the bilateral ground glass, patchy opacities.\\
        \hline
        RadPhi2-Instruct & The endotracheal tube is properly positioned with the tip at the thoracic inlet. A left subclavian central venous line is demonstrated with tip in the mid- superior vena cava. No pneumothorax is present. Bilateral ground glass, patchy opacities are seen. The nasogastric tube is properly positioned in the stomach.\\
        \hline
        RadPhi-3 & There has been interval improvement in the bilateral ground glass, patchy opacities.\\
        \hline
    \end{tabular}
    
    \label{tab:example_radqa_dataset}
\end{table*}

\begin{table*}[!ht]
    \centering
    \caption{Extraction of Tubes and Lines Placement Information}
    \begin{tabular}{|p{50pt}|p{400pt}|}
    \hline
         Prompt& 
Given the below radiology report: 
\BlankLine
PREAMBLE: SINGLE FRONTAL VIEW OF THE CHEST

INDICATION: Respiratory failure.

LAST\_PARAGRAPH: Comparison is made with prior study, \_. Cardiac size is normal. Large right and moderate left pleural effusions are grossly unchanged allowing for differences in positioning of the patient. NG tubes are in the stomach. ET tube is in the standard position. Right peripherally inserted catheter tip is in the right subclavian vein, unchanged.
\BlankLine
Output the full placement description of the devices mentioned in the report.
If no device is mentioned return N/A.\\
         \hline
        Ground Truth &  NG tubes are in the stomach. ET tube is in the standard position. Right peripherally inserted catheter tip is in the right subclavian vein.\\
        \hline
        Phi3-mini-4k-instruct& N/A
Explanation:
The radiology report does not mention any specific devices that are placed within the patient's body. It only mentions the presence of NG (nasogastric) tubes in the stomach and an ET (endotracheal) tube in the standard position, but these are not considered devices that are placed for the purpose of the radiological examination.\\
        \hline
        RadPhi2-Instruct & NG tubes are in the stomach. ET tube is in the standard position. Right peripherally inserted catheter tip is in the right subclavian vein.\\
        \hline
        RadPhi-3 & NG tube is in the stomach. ET tube is in the standard position. Right peripherally inserted catheter tip is in the right subclavian vein.\\

        \hline
    \end{tabular}
    
    \label{tab:example_extraction_of_tubesandlines}
\end{table*}
\clearpage

\section{Hallucination Measurements}\label{sec: hallucination}

\begin{table*}[!ht]
\caption{Align Score Metric for Dataset Creation - Radiology QA Dataset}
\centering 
\footnotesize
\begin{tabular}{|lcc|}
\hline
\textbf{Task} & \textbf{Align Score Mean} & \textbf{Align Score Median}\\
\hline
Radiopaedia QA-Pairs using GPT-4 & $0.89 \pm 0.10$ & 0.92\\
\bottomrule
\multicolumn{3}{p{450pt}}{The alignment score metrics for the Radiology QA dataset with respect to the Radiopaedia articles are very high, indicating that GPT-4 didn't hallucinate while creating the dataset.}
\end{tabular}
\label{tab:Align_score_radiopaedia}
\end{table*}

\begin{table}[!ht]
\centering
\caption{Radiologist Evaluation - Radiology QA Dataset Error Counts}
\begin{tabular}{|lccc|}
\hline
\textbf{Article Count} & \textbf{Hallucination of Facts} & \textbf{Instruction-following Hallucination} & \textbf{Quality Errors}\\
\hline
82&0&1&5\\
\hline
\multicolumn{4}{p{490pt}}{Error counts for the Radiology QA Dataset created by GPT-4. Hallucination of facts refers to errors related to facts in the answers that were not present in the article. Instruction-following hallucinations refer to inadequate adherence to instructions while answering a question. Quality errors refer to question-answer pairs that are not very useful in a clinical setting.}
\end{tabular}
\label{tab:radiologist eval_radiopaedia}
\end{table}

\begin{table}[!ht]
\centering
\caption{Radiologist Evaluation - Temporal Summary Dataset Error Counts}
\begin{tabular}{|lccc|}
\hline
\textbf{No of Temporal Extractions} & \textbf{Hallucination of Facts}  & \textbf{Hallucination of Category} & \textbf{Missed Findings} \\
\hline
50 & 0 & 5 & 7 \\
\hline
\multicolumn{4}{p{500pt}}{Error counts for the Temporal Summary Dataset created by GPT-4. Hallucination of facts refers to errors related to facts in the answers that were not present in the current or prior report. Hallucination of category refers to errors where the change summary category (new, improved, worsened, resolved, etc.) was incorrectly assigned. Missed findings errors are related to the missed change mentions.}
\end{tabular}
\label{tab:radiologist eval_temporal_summary}
\end{table}

\begin{table}[!ht]
\centering
\caption{Radiologist Evaluation - Impression Prediction}
\begin{tabular}{|lcccc|}
\hline
\textbf{No. of Findings} & \textbf{Missed Findings} & \textbf{False Findings} & \textbf{Position Errors} & \textbf{Severity Errors}\\
\hline
149&14&1&2&0\\
\hline
\multicolumn{5}{p{440pt}}{Error counts by category for the impression prediction task. The errors were mostly related to the findings missed in the predictions. The 149 findings were associated with 50 impressions.}
\end{tabular}
\label{tab:radiologist eval}
\end{table}

\begin{table*}[!ht]
    \centering
    \caption{Radiologist Feedback Example - Temporal Change Summary Dataset}
    \begin{tabular}{|p{50pt}|p{400pt}|}
    \hline
         Prompt&  Given a current radiology report and a previous report of the same patient, generate a change summary

Current Report:
                                 FINAL REPORT
 EXAMINATION:  CHEST (PORTABLE AP)
 
 INDICATION:  \_ year old man with ventilator dependence   interval scan     
 ?CXR CHANGES
 
 IMPRESSION: 
 
 Compared to chest radiographs \_ through \_.
 
  Bibasilar atelectasis, severe on the right, moderate on the left, unchanged
 since earlier on \_.  Moderate right pleural effusion is probably
 present.  There is no pulmonary edema although pulmonary vasculature is mildly
 engorged.  No pneumothorax.
 
 Cardiopulmonary support devices in standard placements.

Previous Report:
                                 FINAL REPORT
 EXAMINATION:  CHEST (PORTABLE AP)
 
 INDICATION:  \_ year old man with Right-sided occipital craniotomy for
 evacuation.   interval change
 
 TECHNIQUE:  Single frontal view of the chest
 
 COMPARISON:  \_
 
 IMPRESSION: 
 
 Cardiomediastinal contours are unchanged.  Moderate right pleural effusion has
 increased with associated adjacent atelectasis.  Mild vascular congestion has
 increased.  ET tube is in standard position.  Right subclavian catheter tip is
 in the lower SVC.  NG tube tip is out of view below the diaphragm\\
         \hline
        GPT-4 & Diseases\_Change\_Summary:
	
 \quad New:

 \quad Resolved:

\quad Stable:

  \quad \quad Bibasilar atelectasis, severe on the right, moderate on the left
		Moderate right pleural effusion
		No pulmonary edema
		Pulmonary vasculature is mildly engorged

 \quad Improved:

 \quad Worsened:

 \quad Negatives:

  \quad \quad No pneumothorax

Tubes\_Lines\_Change\_Summary:

 \quad New:

\quad  Removed:

		\quad \quad ET tube
		Right subclavian catheter
		NG tube

 \quad Unchanged:

 \quad Changed:

\quad  Recommendations:
\\
        \hline
        Radiologist Feedback & The devices are not removed, all of them have been captured as `Cardiopulmonary support devices in standard placements'.\\
        \hline
    \end{tabular}
    
    \label{tab:example_temporal_change_summary_error}
\end{table*}

\begin{table*}[!ht]
    \centering
    \caption{Radiologist Feedback Example - Radiology QA Dataset}
    \begin{tabular}{|p{100pt}|p{300pt}|}
    \hline
    Question & What are the different panda signs?\\
    \hline
         Answer Generated by GPT-4 using the  \href{https://radiopaedia.org/articles/panda-sign-disambiguation}{Radiopaedia Article} &  
         1. panda sign of the midbrain, 
         
         2. double panda sign, 
         
         3. panda sign of sarcoidosis, 
         
         4. panda eyes (base of skull fracture)\\
         \hline
        Radiologist Feedback & The Radiopaedia article didn't have enough content to answer this question. The actual answer should contain more details like the below: 
        Panda sign are present in various following pathology and seen under various investigation search:

1. sarcoidosis: under 67 gallium scan denoting uptake of it by lacrimal and parotid gland

2.panda eyes : seen in fracture of anterior of anterior cranial fossa ( dark coloured area around the eyes)

3.double panda sign: t2 mri indicating Wilson dis

4.panda sign of mid brain: t2 mri indication of Wilson dis\\
        \hline
    \end{tabular}
    
    \label{tab:example_radiopaedia_qa_error}
\end{table*}

\begin{table*}[!ht]
    \centering
    \caption{Radiologist Feedback Example - Radiology QA Dataset}
    \begin{tabular}{|p{100pt}|p{300pt}|}
    \hline
    Question & What is the treatment and prognosis for cystic nephromas?\\
    \hline
         Answer Generated by GPT-4 using the         \href{https://radiopaedia.org/articles/cystic-nephroma-2}{Radiopaedia Article} &  
         Radical or partial nephrectomy is usually done, with lymph node excision\\
         \hline
        Radiologist Feedback & The model has answered about treatment but not the prognosis. The Radiopaedia article did not contain the prognosis.\\
        \hline
    \end{tabular}
    
    \label{tab:example2_radiopaedia_qa_error}
\end{table*}

\begin{table*}[!ht]
    \centering
    \caption{Radiologist Feedback Example - Radiology QA Dataset}
    \begin{tabular}{|p{100pt}|p{300pt}|}
    \hline
    Question-Answer Generated by GPT-4 using the         \href{https://radiopaedia.org/articles/hepatic-capsular-retraction-mnemonic-1}{Radiopaedia Article} & [{'question': 'What are the mnemonics for hepatic capsular retraction?', 'answer': 'CT CHIEF, FILTH E, and FITCH are mnemonics for hepatic capsular retraction.'}, {'question': 'What does CT CHIEF stand for?', 'answer': 'C: cholangiocarcinoma (intrahepatic), T: treated hepatic metastasis or hepatocellular carcinoma, C: cirrhosis with confluent hepatic fibrosis, H: hemangioma (especially sclerosed hepatic hemangioma), I: inflammatory pseudotumor, hepatic, E: epithelioid hemangioendothelioma, F: fibrolamellar hepatocellular carcinoma.'}, {'question': 'What does FILTH E stand for?', 'answer': 'F: fibrosis in confluent hepatic fibrosis, I: intrahepatic cholangiocarcinoma, L: lamellar as in fibrolamellar hepatocellular carcinoma, T: treated metastasis, H: hepatocellular carcinoma (uncommon manifestation), E: epithelioid hemangioendothelioma.'}, {'question': 'What does FITCH stand for?', 'answer': 'F: fibrolamellar hepatocellular carcinoma, I: inflammatory pseudotumor, hepatic, T: treated hepatic metastasis or hepatocellular carcinoma, C: cholangiocarcinoma, confluent hepatic fibrosis, H: hemangioma, hemangioendothelioma.'}]\\
         \hline
        Radiologist Feedback & Question-Answer pairs should be more clinically useful, this one is just abbreviations.\\
        \hline
    \end{tabular}
    
    \label{tab:example3_radiopaedia_qa_error}
\end{table*}

\end{document}